\documentclass[conference]{IEEEtran}
\usepackage{times}

\usepackage[numbers,sort]{natbib}
\usepackage[pdftex]{graphicx}
\usepackage{amsmath} \usepackage{amssymb}  \usepackage{amsfonts,bm}
\usepackage{siunitx}
\usepackage{xspace}
\usepackage[export]{adjustbox}
\usepackage{booktabs, array}
\usepackage{subcaption,dcolumn}

\usepackage{blindtext} 
\usepackage{soul}
\usepackage{lib} 
\usepackage{alias} 
\usepackage{multicol,multirow} 
\usepackage{tabularx}
\newcolumntype{C}{>{\centering\arraybackslash}X}
\NewExpandableDocumentCommand\mcc{O{1}m}{\multicolumn{#1}{c}{#2}}
\usepackage{algpseudocode}
\usepackage{algorithm}
\usepackage{cellspace}
\usepackage{wrapfig}
\usepackage[dvipsnames]{xcolor}
\newcolumntype{d}[1]{D..{#1}}

\usepackage[labelfont=bf,size=footnotesize]{caption}
\usepackage[breaklinks=true,letterpaper=true,colorlinks=true,citecolor=blue,bookmarks=true]{hyperref}

\definecolor{olivegreen}{RGB}{85,107,47}
\definecolor{darkgreen}{RGB}{0,100,0}

\begin{document}
\title{Diffusion Meets DAgger: \\ Supercharging Eye-in-hand Imitation Learning}

\author{\authorblockN{Xiaoyu Zhang \quad
Matthew Chang \quad
Pranav Kumar \quad 
Saurabh Gupta}
\authorblockA{
University of Illinois at Urbana-Champaign\\
\texttt{\{zhang401,mc48,pranav9,saurabhg\}@illinois.edu}\\
\url{https://sites.google.com/view/diffusion-meets-dagger}}
}

\maketitle

\begin{abstract}
A common failure mode for policies trained with imitation is compounding execution errors at test time. When the learned policy encounters states that are not present in the expert demonstrations, the policy fails, leading to degenerate behavior.
The Dataset Aggregation, or DAgger approach to this problem simply collects more data to cover these failure states. However, in practice, this is often prohibitively expensive.
In this work, we propose Diffusion Meets DAgger (\name), a method that reaps
the benefits of DAgger but without the cost, for eye-in-hand imitation learning
problems.
Instead of \textit{collecting} new samples to cover out-of-distribution states, \name uses recent advances in diffusion models to \textit{synthesize} these samples. This leads to robust performance from few demonstrations. 
We compare \name against behavior cloning baseline across four tasks: pushing, stacking, pouring, 
and hanging a shirt.
In pushing, \name achieves
80\% success rate with as few as 8 expert demonstrations,
where naive behavior cloning reaches only 20\%. 
In stacking, \name succeeds on average 92\% of the time across 5 cups, versus 40\% for BC. 
When pouring coffee beans, \name transfers to another cup successfully 80\% of the time. 
Finally, \name attains 90\% success rate for hanging shirt on a clothing rack.

 \end{abstract}

\IEEEpeerreviewmaketitle

\section{Introduction}
\seclabel{intro}
Imitation learning is an effective way to train robots for new tasks.
However, even when testing policies in environments similar to those used for
training, imitation learning suffers from the well-known {\it Compounding
Execution Errors problem}: small errors made by the policy lead to out-of-distribution states, causing the robot to make even bigger
errors (compounding errors)~\cite{ross2011reduction}.

One solution to this problem is via manual collection of
expert-labeled data on states visited by the learner, a strategy popularly
known as Dataset Aggregation or DAgger~\cite{ross2011reduction}. 
However, DAgger~\cite{ross2011reduction} 
is challenging to put into practice: it requires an expert operator to supervise the robot during execution and guide it to recover from failures.
Many alternatives have been proposed, \eg ~\cite{laskey2016shiv, laskey2017dart}, but they
all require collecting more expert data. In this paper, we pursue an alternate
paradigm: {\it automatically generating observations and action labels for
out-of-distribution states}. By replacing data collection with data creation,
we improve the sample efficiency of imitation learning. In fact, data creation
was Pomerleau's original solution to this
problem~\cite{pomerleau1991efficient}. We revisit, improve, and automate his
solution from 30 years ago using modern data-driven image generation
methods.

\begin{figure}
\includegraphics[width=\linewidth]{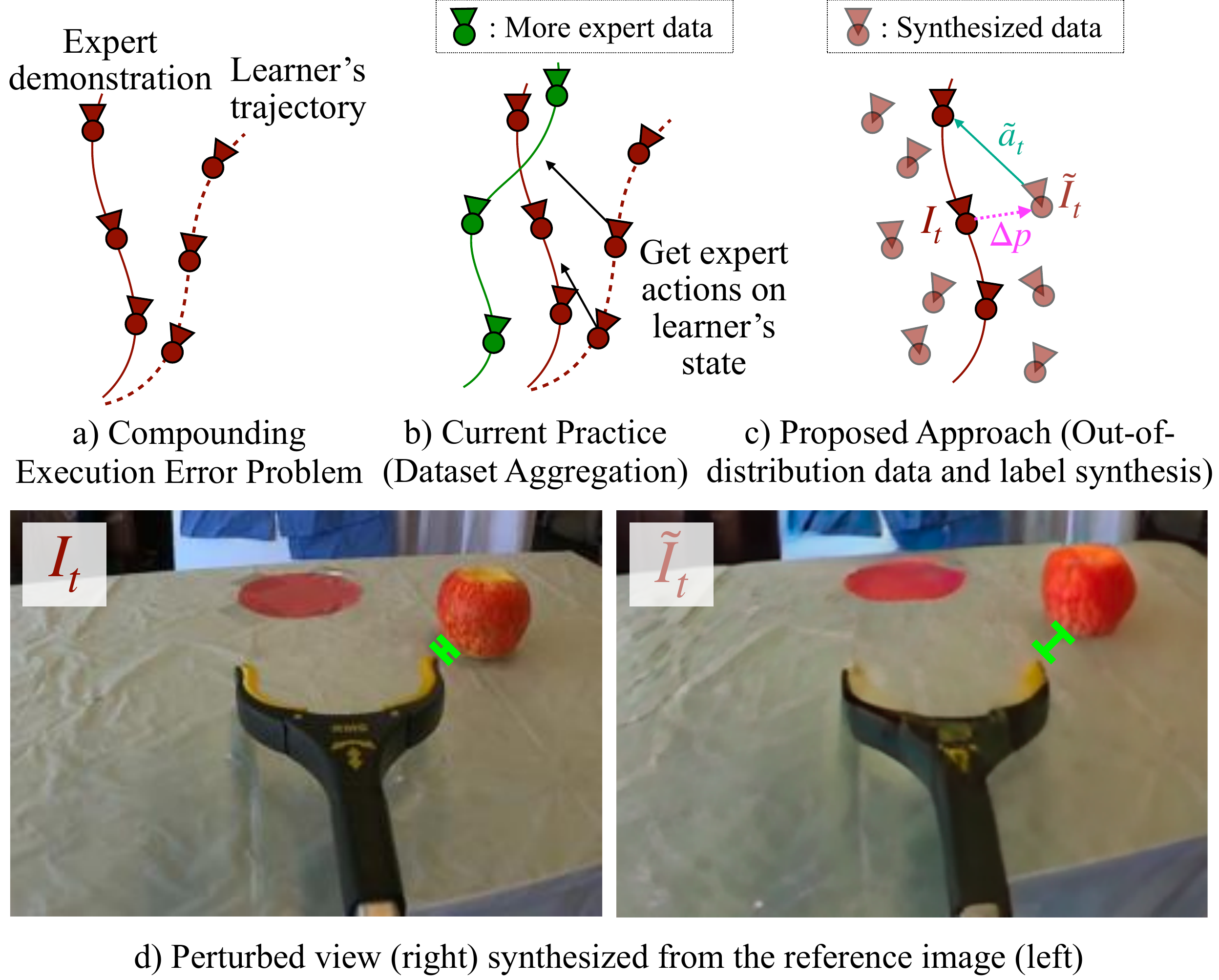} \\
\caption{\textbf{Eye-in-hand Imitation learning with \name:} 
A common failure mode in an imitation learning setting is the problem of poor generalization due to compounding execution errors at test time as shown in (a). This can be solved by collecting more expert data to cover these off-trajectory states as shown in (b) however, this is an expensive process. 
Our proposed approach is to \textit{synthesize} data instead of \textit{collecting} it (c). 
{\color{magenta}Magenta arrow} represents small perturbation ($\Delta p$) to the trajectory.
{\color{BlueGreen}Cyan arrow} represents label ($\tilde{a}_t$) for this out-of-distribution observation.
We use a state-of-the-art diffusion model to take images $I_t$ from expert demonstrations (d, left) and generate realistic off-trajectory images $\tilde{I}_t$ (d, right). 
Note the distance between the grabber and the apple denoted by the green line. This synthetic data augments expert demonstrations for policy learning, leading to more robust policies.}
\figlabel{teaser}
\end{figure}

We target imitation learning problems in the context of eye-in-hand setups (\ie setups where images come from a camera mounted on the robot hand), which are becoming increasingly more popular~\cite{song2020grasping, zhao2023learning, young2021visual}.  
Given a behavior cloning trajectory $\tau$ collected from an
expert, we design generative models to synthesize off-distribution states and compute corresponding action labels. 
For example, in the object pushing application shown in \figref{teaser}, this corresponds to
generating off-center views of the object from the good expert trajectory
$\tau$.  Specifically, we learn a function $f(I_t, \Delta p)$ that
synthesizes the observation $\tilde{I}_t$ at a small perturbation $\Delta p$
to the trajectory at time step $t$. 
$\Delta p$ lets us
compute the ground truth action label $\tilde{a}_t$ for this out-of-distribution
observation.
We augment the
expert demonstration $\tau$ with multiple off-distribution samples $(\tilde{I}_t,
\tilde{a}_t)$ along the trajectory.

As we have access to the entire trajectory, one option is to realize $f(I_t, \Delta p)$ using NeRFs~\cite{mildenhall2021nerf, zhou2023nerf}. While NeRFs work very well for
view synthesis in computer vision, we find they are not suitable for the synthesis task at hand because of the
inevitable deformations in the scene as the gripper manipulates the objects. 
We thus switch to using diffusion models, and this change leads to high-quality image synthesis even when the scene deforms during manipulation. 
Computing action labels for these samples 
present yet another challenge (\figref{overshooting}). 
The labels should align with progress towards the goal. 
We thus investigate different schemes for
sampling $\Delta p$'s and computing action labels $\tilde{a}_t$. 

We present experiments that evaluate the aforementioned design choices
in developing a data creation framework to supercharge eye-in-hand imitation learning.
The framework is tested in four settings: non-prehensile pushing,
stacking, pouring,
and hanging a shirt. 
Across all tasks, we see a sizeable improvement over
vanilla behavior cloning, demonstrating the effectiveness of our framework Diffusion Meets DAgger (DMD).
For the task of non-prehensile pushing, \name achieves 80\% success rate given only 8 expert demonstrations, while behavior cloning (BC) reaches only 20\%. It also outperforms a NeRF-based augmentation method  
\cite{zhou2023nerf}
by 50\%. 
For the task of stacking cups on a box, DMD achieves a success rate of 93\% on
average across three training cups, and 85\% success for two unseen cups. In
comparison, BC only reaches 36\% and 45\% success rates, respectively.
For the task of pouring coffee beans into a cup, DMD succeeds 80\% of the time,
while BC falls short at 30\%. Finally, DMD achieves a success rate of 90\% for the
  hanging shirt task, improving upon the 20\% success rate for BC.

\section{Related Work}
\seclabel{related}

\subsection{Imitation Learning}
Behavior cloning (BC), or training models to mimic expert behavior, has been a
popular strategy to train robots over the last many
decades~\cite{pomerleau1988alvinn, pomerleau1991efficient, schaal1996learning},
and has witnessed renewed interest in recent times~\cite{jang2022bc, zhu2018reinforcement}.
\cite{ross2011reduction} theoretically and empirically demonstrate the
compounding execution error problem (small errors in the learner's predictions
steer the agent into out-of-distribution states causing the learner to make
even bigger errors) and propose a Dataset Aggregation (DAgger) strategy to
mitigate it. 

Over the years, researchers have sought to improve the basic BC
and DAgger recipe in several ways~\cite{hussein2017imitation,
argall2009survey}.  \cite{song2020grasping, young2021visual, 7358076,
zhao2023learning, shafiullah2023bringing} devise ways to ease and scale-up
expert data collection, while \cite{zhang2018deep} design ways to collect
demonstrations in virtual reality.  \cite{laskey2016shiv, laskey2017dart}
devise ways to simplify DAgger data collection.  \cite{chi2023diffusionpolicy,
shafiullah2022behavior} model the multiple modes in expert demonstrations,
while \cite{pari2021surprising} employ a non-parametric approach. Researchers
have also shown the effectiveness of pre-trained representations for
imitation learning~\cite{pari2021surprising, xiao2022masked}. 
\cite{zhou2023nerf, chen2023genaug, mandi2022cacti, yu2023scaling}
employ image and view synthesis models to synthetically augment data. 
Complementary to our work, \cite{chen2024simple, mandi2022cacti, yu2023scaling}
focus on generalization across objects via {\it semantic} data augmentation by
adding, deleting and editing objects in the scene.

~\cite{zhou2023nerf, ke2021grasping, florence2019self, park2022robust} 
pursue synthetic data augmentation to address the compounding execution error
problem. 
~\cite{ke2021grasping, florence2019self, park2022robust} 
conduct these augmentation in
low dimensional state spaces. 
\cite{park2022robust} 
develops a principled way
to sample these augmentations.
Closest to our work, \citet{zhou2023nerf}'s SPARTN
model train NeRF's on demonstration data to synthesize out-of-distribution
views. NeRF assumes a static scene. Thus, SPARTN can't reliably synthesize
images when the scene undergoes deformation upon interaction of the robot with
the environment. Our use of diffusion models to synthesize images gets around
this issue and experimental comparison to SPARTN demonstrate the effectiveness
of our design choice.

\subsection{Image and View Synthesis Models}
Recent years have witnessed large progress in image
generation~\cite{goodfellow2014generative, ho2020denoising, rombach2021high, sohl2015deep, song2019generative}. This has led to a
number of applications: text-conditioned image
generation~\cite{rombach2021high, balaji2022ediffi, gu2022vector,
saharia2022photorealistic, ramesh2022hierarchical, zhang2023adding}, image and
video inpainting~\cite{lugmayr2022repaint, chang2023look, meng2021sdedit},
image to image translation~\cite{isola2017image} and generation of novel views for objects
and scenes from one or a few images~\cite{watson2022novel, tseng2023consistent,
yu2023long, liu2023zero, kwak2023vivid}.  While diffusion models have proven
effective at image synthesis tasks, Neural Radiance Fields
(NeRFs)~\cite{mildenhall2021nerf} excel at view interpolation tasks. Given
multiple (20 -- 100) images of a static scene, vanilla NeRFs can effectively
interpolate among the views to generate photo-realistic renders.  Researchers
have pursued NeRF extensions that enable view synthesis from few
images~\cite{yu2021pixelnerf, gu2023nerfdiff, blukis2022neural, yang2023freenerf} and even model
deforming scenes~\cite{su2021nerf, cao2023hexplane}. In general, NeRFs are more effective at interpolation
problems, while diffusion models excel at problems involving extrapolation (\eg
synthesizing content not seen at all) or where exactly modeling changes in 3D
is hard (\eg deforming scenes). As our work involves speculating how the scene
would like if the gripper were at a different location, we adopt a diffusion
based methods. We show comparison against a NeRF based method as well.

\subsection{Diffusion Models and Neural Fields for Robot Learning} Effective
generative models and neural field based representations have been used in
robotics in other ways than for data augmentation.  Researchers have
used diffusion models for representing policies~\cite{li2023hierarchical,
ada2024diffusion, li2023crossway, chi2023diffusionpolicy, wang2022diffusion}, generating goals or subgoals
\cite{kapelyukh2023dall,black2023zero,chen2024simple}, offline reinforcement
learning~\cite{ajay2022conditional, wang2022diffusion, lu2023synthetic},
planning~\cite{liang2023adaptdiffuser, janner2022planning,
xian2023chaineddiffuser, liang2023skilldiffuser}, behaviorally diverse policy
generation~\cite{hegde2023generating}, predicting
affordances~\cite{ye2023affordance}, and skill acquisition from task or play
data~\cite{chen2023playfusion}.
Neural fields have been used to represent scenes for manipulation~\cite{li20223d, wang2023d} and navigation~\cite{adamkiewicz2022vision}.

\begin{figure*}[t]
    \includegraphics[width=1.0\textwidth]{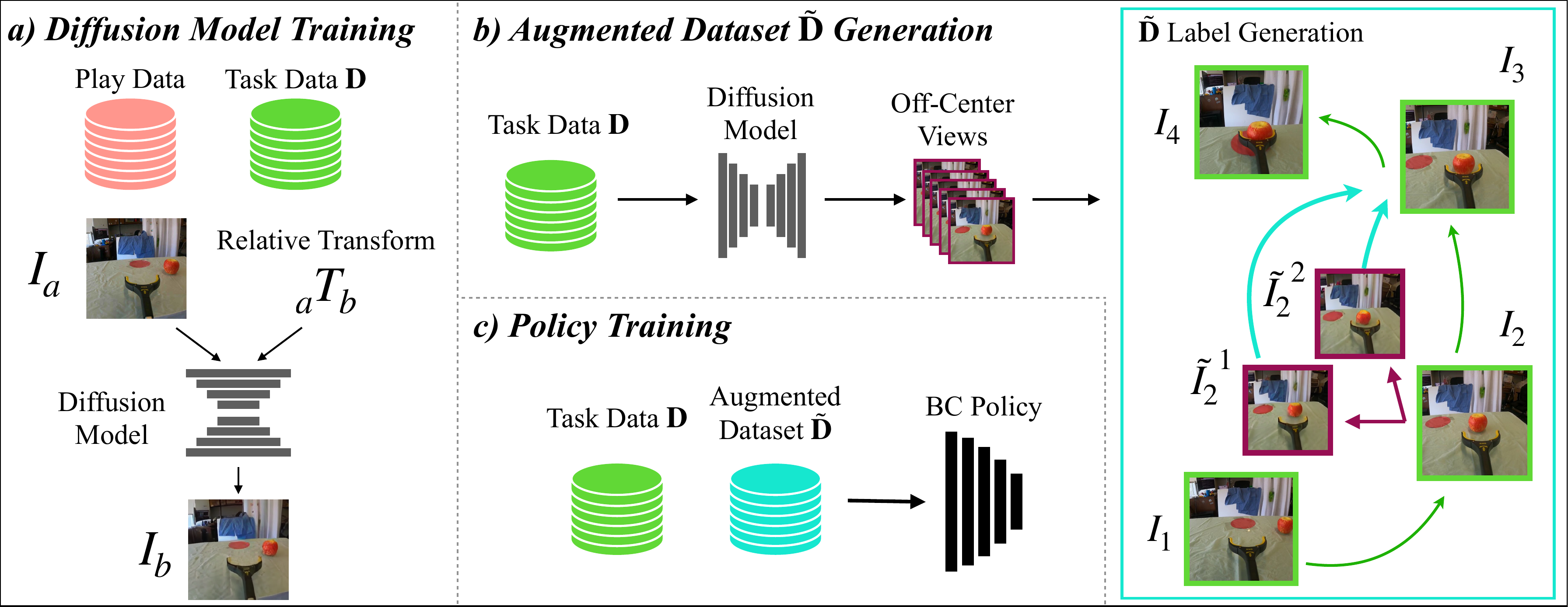}
\caption{{\bf \name System Overview:} 
    Our system operates in three stages. 
    {\bf a)} A diffusion model is trained, using task and play data, to synthesize novel views relative to a given image. 
    {\bf b)} This diffusion model is used to generate an augmenting dataset that contains off-trajectory views {\color{purple}($\tilde{I_2^1}$, $\tilde{I_2^2}$)} from expert demonstrations. 
    Labels for these views ({\color{BlueGreen}cyan arrows}) are constructed such that off-trajectory views will still converge towards task success (right).
    {\color{darkgreen} Images with a green border} are from trajectories in the original task dataset. 
    {\color{purple}Purple-outlined images} are diffusion-generated augmenting samples.
    {\bf c)} The original task data and augmenting dataset are combined for policy learning.
    }
    \figlabel{arch}
    \end{figure*}

\section{Approach}
\seclabel{approach}

\subsection{Overview}
Given task data $\Dtask$, imitation learning learns a task policy $\pi$. 
The task data comprises a set of trajectories $\tau_i$, each consisting of a sequence of image action pairs, $(I_t, a_t)$.
Policy $\pi$
is trained using supervised learning to regress action $a_t$ from images $I_t$. 
In this paper, we consider manipulation of objects using an eye-in-hand camera, where input images $I_t$ correspond to views from a wrist camera, and the actions $a_t$ are the relative end-effector
poses between consecutive time steps.

When trained with few demonstrations, $\pi$ exhibits poor online
performance upon encountering states not covered by the expert demonstrations.
To address this issue, as shown in \figref{arch}, our approach generates an
augmented dataset $\Daug$ and trains the policy jointly on $\Daug \cup \Dtask$.
Augmented samples are specifically generated to be out-of-distribution from
$\Dtask$, thus helping the policy to generalize. $\Daug$ is
generated through a conditional diffusion model $f$ that synthesizes a $\Delta p$-perturbed view $\tilde{I}_t$ of an image $I_t$ from a given trajectory
$\tau$: $\tilde{I}_t = f(I_t, \Delta  p)$. 
We use action labels in the trajectory $\tau$ to compute the action label 
$\tilde{a}_t$  for this perturbed view.  The design of
the conditional diffusion model is described in \secref{nvs}, and the procedure for sampling augmenting images and computing action labels is detailed in \secref{gen}.

\subsection{Diffusion Model for Synthesizing $\Delta  p$ Perturbed Views}
\seclabel{nvs}
The function $f(I_t, \Delta  p)$ is realized using a conditional diffusion model
that is pretrained on large Internet-scale data. Specifically, we adopt the
recent work from \citet{yu2023long} and finetune it on data from $\Dtask$. 
As shown in \figref{arch}(a), the model
produces image $I_b$ by conditioning on a reference image $I_{a}$ taken by
camera $a$ and a transformation matrix ${}_{a}T_{b}$ that maps points in camera $b$'s frame to camera $a$'s frame. By representing the desired $\Delta p$ as the transformation between two cameras (${}_aT_b = \Delta p)$, and using images from $\mathbf{D}$ as the reference images ($I_a \sim \mathbf{D}$), the learned model can generate the desired perturbed views for augmenting policy learning.

\subsubsection{Model Architecture}

A diffusion model is an iterative denoiser:
given a noisy image $x_t$, diffusion timestep $t$, it is
trained to predict the noise $\epsilon$ added to the image: \[\mathcal{L} =
\|\epsilon - \epsilon_\theta(x_t,t)\|^2_2.\]

It is typically realized via a U-Net~\cite{ronneberger2015u}. Our application
requires {\it conditional image generation} where the conditions are source
image $I_a$ and transformation ${}_{a}T_{b}$. Thus, by using $I_b$ as the diffusion target, and conditioning on $I_a$ and ${}_{a}T_{b}$, the model $\epsilon_\theta$ learns to denoise new views of the scene in $I_a$ based on the transformation ${}_{a}T_{b}$. Finally, rather than directly denoising in the
high-resolution pixel space, the denoising is done in the latent space of a
VQ-GAN autoencoder, $E$~\cite{esser2020taming,rombach2021high}. This gives the final training objective of: 
\[\mathcal{L} = ||\epsilon - \epsilon_\theta(x^b_t, E(I_a), {}_{a}T_{b},t)||\quad \text{where}\quad x^b_0 = E(I_b).\]
Following \cite{yu2023long}, the pose
conditioning information ${}_aT_b$ is injected into the U-Net via cross-attention as
shown in \figref{diffusion-arch}.

\begin{figure}[t]
    \includegraphics[width=\linewidth]{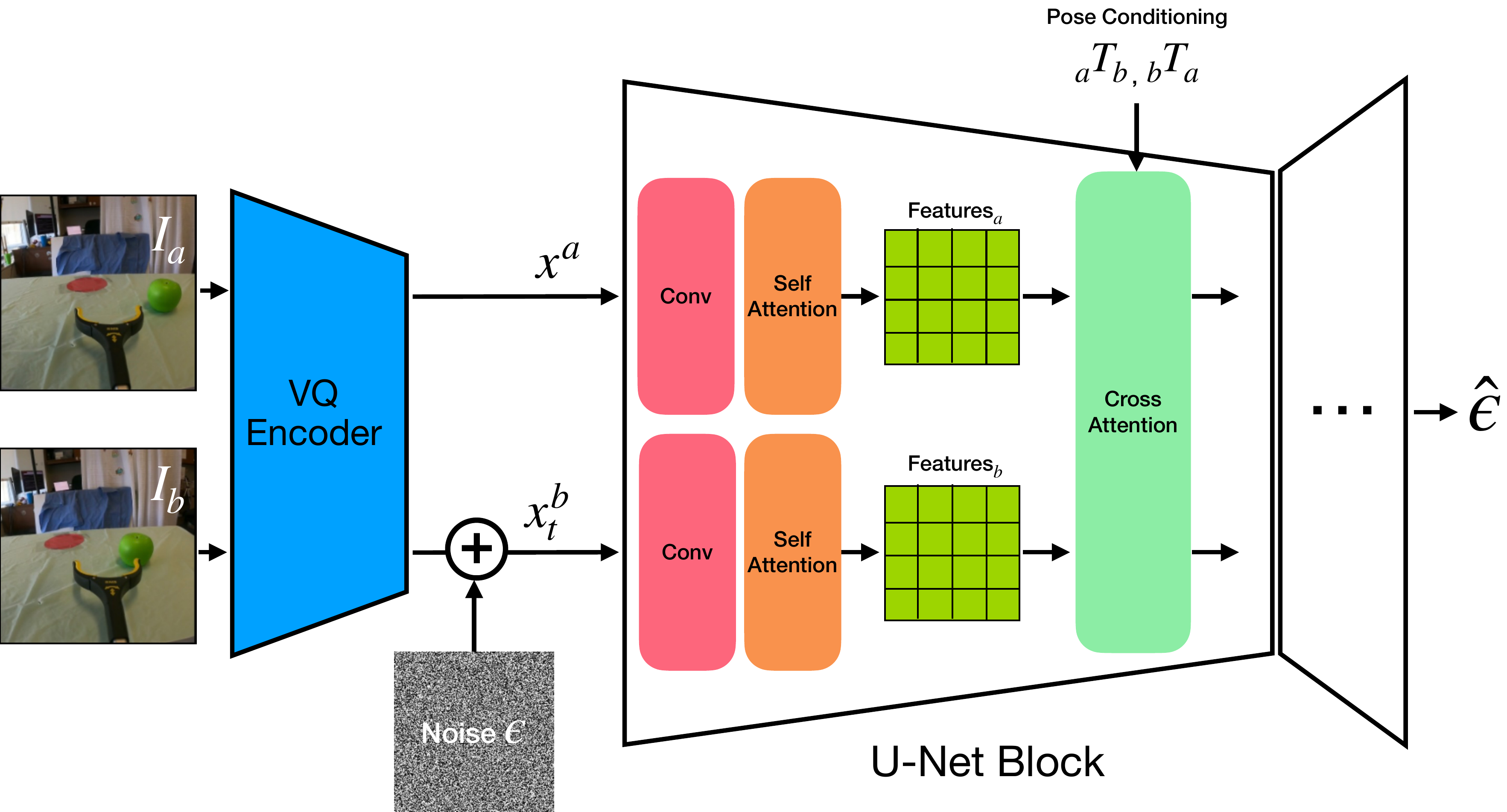}
    \caption{{\bf \name Architecture:} We use the architecture introduced in \cite{yu2023long}, a U-Net diffusion model with blocks composed of convolution, self-attention, and cross attention layers. The conditioning image $I_a$, and noised target image $I_b$ are processed in parallel except at cross-attention layers. The pose conditioning information is injected at cross-attention layers.}
    \figlabel{diffusion-arch}
    \end{figure} 
\subsubsection{Model Training} 
Training the model requires access to triples: $(I_a, I_b, {}_{a}T_{b})$. We
process data from $\Dtask$ to generate these triples.  Let each trajectory
consist of images $I_1,\dots,I_N$.  
We use structure from motion (SfM) algorithms~\cite{hartley2003multiple, thrun2002probabilistic, schoenberger2016sfm, schoenberger2016mvs, ORBSLAM3_TRO,chi2024universal}
to extract poses for the
images in the trajectory $\tau$. This associates each
image $I_t$ in the trajectory with the (arbitrary) world frame ${}_{t}T_{w}$.
We can then compute relative pose between arbitrary images $I_a$ and
$I_b$ from a trajectory via ${}_{a}T_b = {}_{a}T_w {}_{w}T_b$. We sample random
pair of images from the trajectory to produce triples $(I_a, I_b, {}_aT_{b})$
that are used to train the model $\epsilon_\theta$.

\begin{figure}
    \centering
\includegraphics[width=1.0\linewidth]{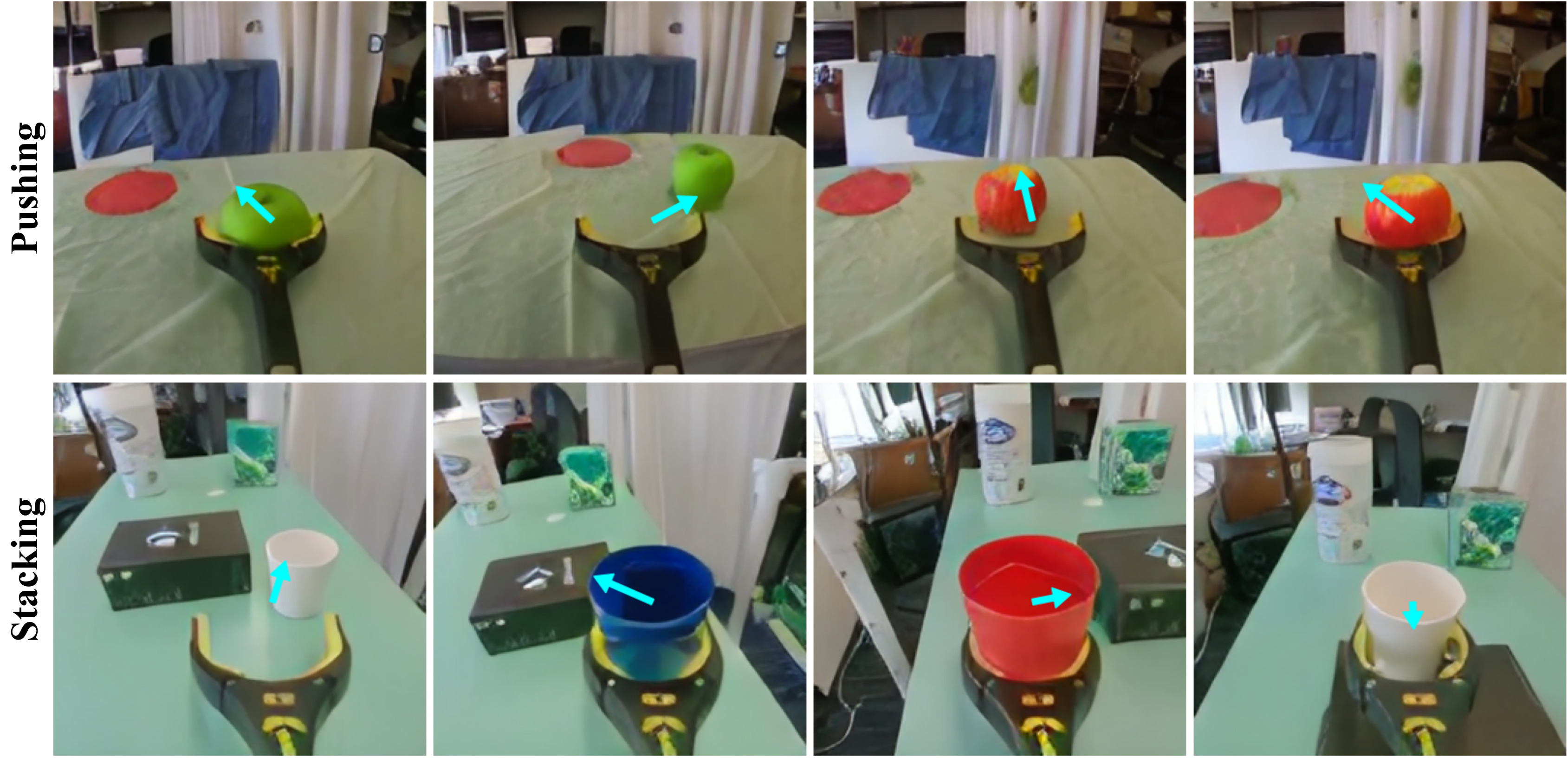}
    \caption{\textbf{Training Examples from the Diffusion Model and Computed Labels:} 
    We visualize generated examples, $\tilde{I}$, used to train our policies along with the computed action label (the arrow is a projection of the 3D action into the 2D image plane: the arrow pointing up means move the gripper forward, pointing to the right means move it right). 
The first row shows augmenting samples for the pushing task, while the second row shows those for the stacking task.}
\figlabel{diffusion-samples-actions}
\end{figure}

As $\Dtask$ contains much fewer data than is typically required to train a
diffusion models from scratch, we finetune the model from \citet{yu2023long}.
Finetuning with around 50 trajectories leads to realistic novel view synthesis
for our tasks as shown in \figref{diffusion-samples}.  We only finetune the
diffusion model weights. The pre-trained VQ-GAN codebooks is kept fixed.

The diffusion model can be finetuned on task data $\Dtask$, task-agnostic play
data~\cite{lynch2020learning, young2022playful}, or even combinations of task
and play data. Play data in our setups involves moving the grabber within the
workspace and randomly interacting with the task objects 
without emphasizing task completion.  
Access to random interactions in play data may make it easier
for the diffusion model to synthesize diverse perturbations. We experiment with
these different data sources for training the diffusion model
(\secref{push_utility_play} and \secref{pour-online-exp}).

\subsection{Generating Out-of-Distribution Images and Labels}
\seclabel{gen}

\subsubsection{Sampling Out-of-distribution Images}

With the noise estimation model $\epsilon_\theta$ trained, the image generation function $f(I_t,\Delta p)$ is realized by simply letting $I_a = I_t$, letting ${}_aT_b$ represent the desired $\Delta p$, and sampling from the conditional diffusion model using well-established sampling strategies \cite{song2020score,ho2020denoising,rombach2021high}.

We use $f(I_t, \Delta p)$ to generate out-of-distribution images with a very simple strategy for sampling $\Delta p$. 
For an image $I_t$ from trajectory $\tau$, we sample vectors in a random direction, with magnitudes drawn uniformly from a pre-defined range for each task.
If the SfM reconstruction recovers the real-world scale, this range is set to $[2cm, 4cm]$; otherwise, it is $[0.2s, s]$, where $s$ is the largest displacement between adjacent frames in $\tau$.
$\Delta p$ simply corresponds to moving along this randomly chosen vector.

\begin{figure}
\includegraphics[width=\linewidth]{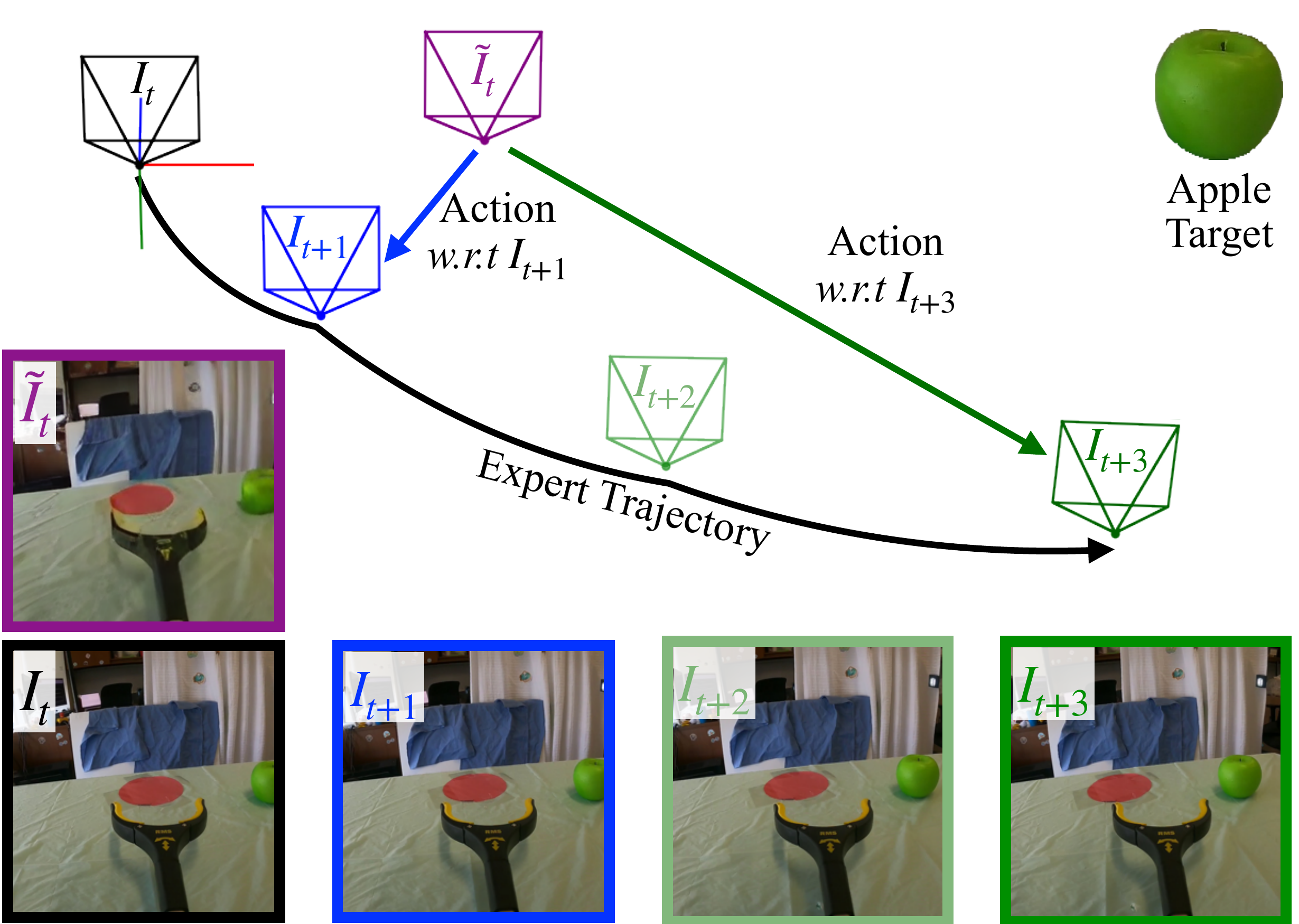}
\caption{{\bf The Overshooting Problem:} 
When the generated image {\color{purple} $\tilde{I}_t$} exceeds {\color{blue} $I_{t+1}$}, the inferred action for $\tilde{I}_t$ may direct the agent away from task success. We refer to this as the overshooting problem. 
At time step $t+1$, the view $I_{t+1}$ has moved to the lower right of $I_t$. 
However, the synthesized sample $\tilde{I}_t$ has moved even further to the right than $I_{t+1}$, but not beyond $I_{t+3}$.
{\color{blue} Blue arrow} represents action label for $\tilde{I}_t$ computed using $I_{t+1}$ as the target; {\color{darkgreen} green arrow} represents action label computed using $I_{t+3}$ as the target. 
Since $\tilde{I}_t$ has \textit{overshot} $I_{t+1}$, an action taken with $I_{t+1}$ as the next intended target moves backward, away from the apple, and this labeling is not desirable. 
Computing the action with respect to a farther image, say $I_{t+3}$, does not have this issue.}
\figlabel{overshooting}
\end{figure} 
\subsubsection{Labeling Generated Images}
\seclabel{label-generate-image}

For each image $\tilde{I}_t = f(I_t,\Delta p)$ generated from an original image $I_t$, we use
$I_{t+k}$ as the target for generating the action label. 
The action label for $\tilde{I}_t$ is simply the action that conveys the agent from the pose depicted in $\tilde{I}_t$ to the pose in $I_{t+k}$. 

Using SfM, we obtain camera pose information ${}_{t}T_w$ for time step $t$ and ${}_{t+k}T_w$ for time step $t+k$.
The synthesized view $\tilde{I}_t$ can be treated as an image taken by a virtual camera whose pose is represented as ${}_{t}T_{\tilde{t}}$.
The diffusion model synthesizes perturbed view conditioned on $I_t$ and ${}_{t}T_{\tilde{t}}\;$, or
$\tilde{I}_t = f(I_t, {}_{t}T_{\tilde{t}})$
(\secref{nvs}).
The action label $\tilde{a}_t$ for $\tilde{I}_t$ is then computed as 
$\tilde{a}_t = {}_{\tilde{t}}T_{t+k} \; = \; {}_{\tilde{t}}T_{t} \, {}_{t}T_w \,({}_{t+k}T_w)^{-1} $.
Examples of $(\tilde{I}_t, \tilde{a}_t)$ are shown in \figref{diffusion-samples-actions}.

One might wonder why we don't just use $k=1$?
When sampling $\Delta p$, we assume that the perturbed observation $\tilde{I}_t$ is within a local region around $I_t$.
However,
some generated samples can move past the target image $I_{t+1}$ (described in \figref{overshooting}).
This causes conflicting supervision, as the computed action does not make progress toward completing the task. Using a
large $k$ guards against this, leading to stronger performance downstream, as our experiments verify (\secref{overshoot-sec}).

\begin{figure*}\includegraphics[width=\linewidth]{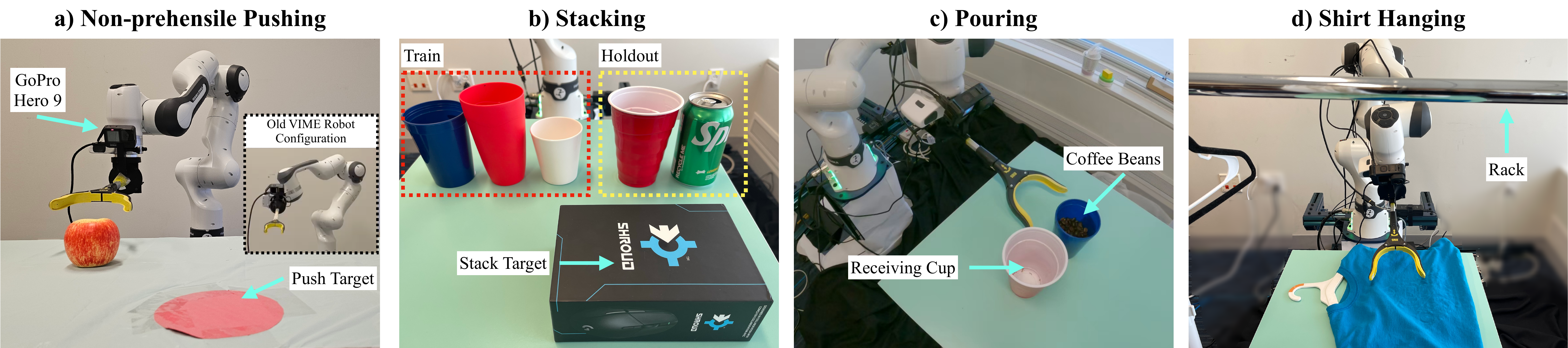}
\caption{{\bf \name Robotic Experiments:} 
There are four tasks in total: pushing apple to target location (\secref{exp-push}),
stacking five different cups on box (\secref{online-stack}),
pouring coffee beans into a cup (\secref{online-pour}),
and hanging shirt on a rack (\secref{online-hanger}).
We conduct our experiments on a Franka Research 3 robot with a wrist-mounted GoPro Hero 9. 
We modified VIME's~\cite{young2021visual} grabber mount for Franka, allowing the robot to reach end-effector poses without reaching joint limits.
}
\figlabel{old-new-robot-config}
\end{figure*} 
\section{Experiments}
\seclabel{experiments}

We test \name across four tasks: non-prehensile pushing, stacking, pouring, and hanging a shirt. The task setups are shown in \figref{old-new-robot-config}. 
Together these tasks test \name in different settings: 
3DoF action space (pushing, stacking), 
6DoF action spaces (pouring, hanging a shirt),
generalization to new objects (stacking), 
precision in reaching goal locations (pouring, hanging a shirt).

On the pushing task, we present visual comparisons to NeRF-based synthesis approach SPARTN~\cite{zhou2023nerf} in
\secref{vis-gen} and in-depth quantitative analysis (ablation of design choices, offline evaluations) in \secref{push-offline}.  
We then show a variety of online comparisons for the
pushing task in \secref{online-exp}.
Moreover, we compare \name to behavior cloning for stacking (\secref{online-stack}),
pouring (\secref{online-pour}), and hanging a shirt (\secref{online-hanger}). 
Finally, we test whether \name improves generalization to novel objects and
environment when provided with a diverse task dataset, as described in
\secref{cup-arrangement}. We adopt randomized A/B testing when making
comparisons between different methods to minimize the effects of
unknown environmental factors.

\begin{figure*}[t]
\includegraphics[width=1.0\linewidth]{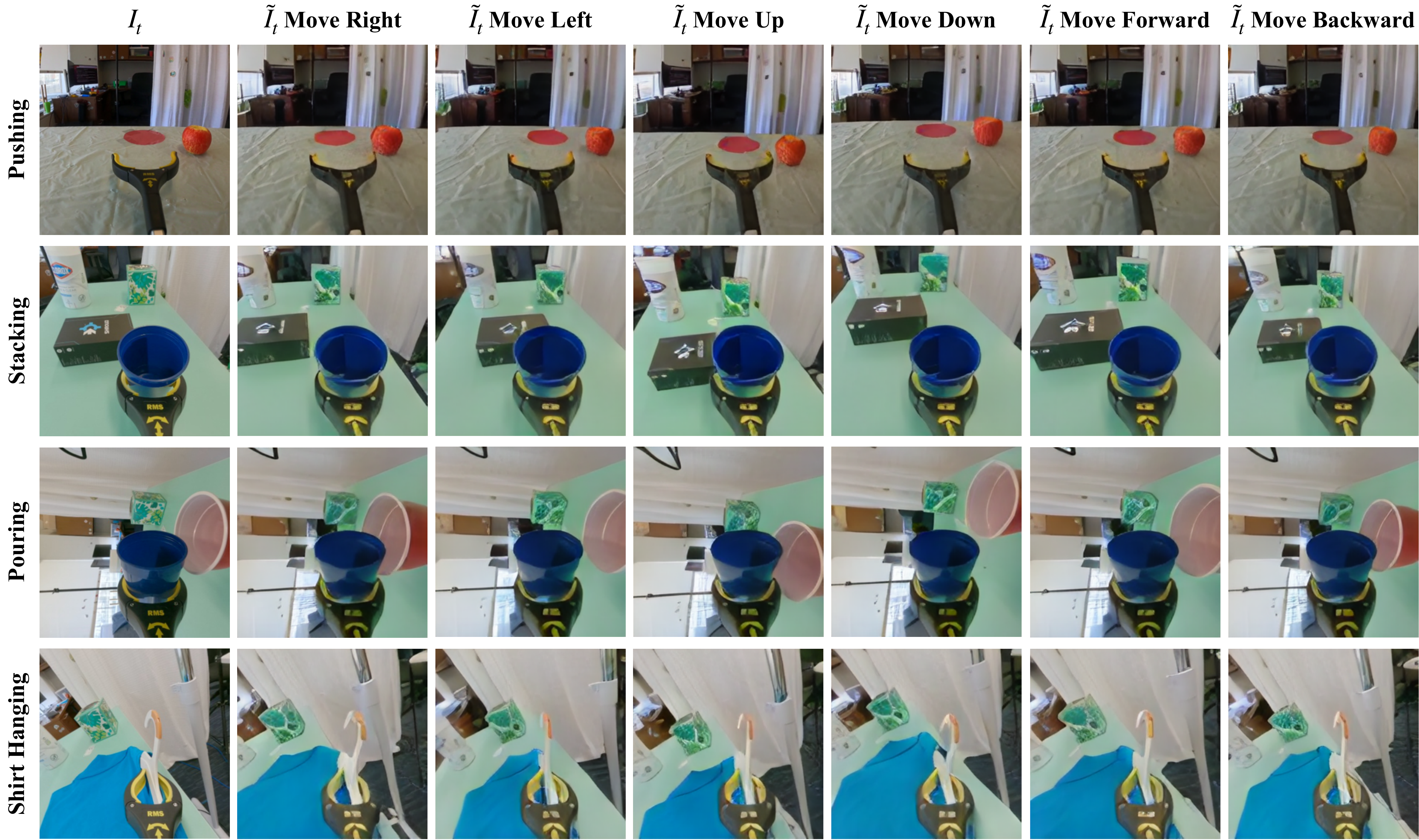}
    \caption{{\bf Visualization of Perturbed Images Generated by Diffusion Model for Augmenting Policy Learning:}  In each row, we show $\tilde{I_t}$ generated from $I_t$ through different camera translations for the tasks of pushing, stacking, pouring, and hanging a shirt.}
    \figlabel{diffusion-samples}
    \end{figure*}

\subsection{\textbf{\textit{Non-prehensile Pushing}}}
\seclabel{exp-push}

\subsubsection{\bf Experiment Setup}
\paragraph{Task, Observation Space, Action Space} 
We use the non-prehensile pushing task from VIME~\cite{young2021visual}, which involves reaching a target object and pushing it to the target location marked by a red circle.
Following \citet{young2021visual}, we use a grabber as the robot end-effector as shown in \figref{old-new-robot-config}. 
This allows for easy collection of expert demonstrations using a GoPro mounted on a grabber. 
Due to the difference in joint limits between the XArm (used by VIME) and the Franka Research 3 robot, we redesign the attachment between the grabber and the robot flange (\figref{old-new-robot-config}(a)). 
In this way, the robot can execute planar push in the typical top-down configuration, extending the workspace to a wider range of the table. 
Observations come from the eye-in-hand GoPro camera. 
For pushing, we use the action space used in VIME's~\cite{young2021visual} publicly available code: relative 3D translations of the end-effector.

\paragraph{Task and Play Data} 
We collect task and play data using a GoPro mounted on a grabber.
The task data include 74 demonstration, and the play data include 36 trajectories.
For task data, an expert pushes the object to a target location.
For play data, we move the grabber around the apple and push it in different directions. 
From the GoPro videos, we obtain image sequences and use structure from motion~\cite{hartley2003multiple, thrun2002probabilistic} to extract camera poses ${}_tT_w$. 
Action $a_t$ is computed by computing the relative translation between $I_{t+1}$ and $I_{t}$ in $I_t$'s camera frame. 
Since COLMAP only gives reconstructions up to an unknown scale factor, we normalize the action to unit length and interpret it as just a direction.

\paragraph{Policy Architecture and Execution} 
We adopt the architecture for the task policy $\pi$ from VIME~\cite{young2021visual} but replace the AlexNet~\cite{10.5555/2999134.2999257} backbone with an ImageNet pre-trained ResNet-18 backbone~\cite{he2015deep}. The policy consists of the ResNet backbone followed by 3 MLPs that predict the action. This task policy accepts a center cropped image from the GoPro camera and outputs the direction in which the camera should move. The policy is trained by L1 regression to the (unit length) actions using the Adam optimizer with a learning rate of 1e-4. Actions are executed on the robot by commanding the robot to go 1cm in the predicted direction.

\paragraph{Baselines} 
We use vanilla behavior cloning on the expert data as the baseline as done in past work~\cite{young2021visual}, we refer to this as BC. We also compare to SPARTN~\cite{zhou2023nerf} and evaluate the various design choices. 
Since there is no publicly available code from SPARTN, we replicate their procedure using NerfStudio~\cite{tancik2023nerfstudio}.

\subsubsection{\bf Visual Comparison of Generated Images}
\seclabel{vis-gen}
\figref{diffusion-samples} shows the high visual quality of samples generated by the diffusion model.
The model is able to faithfully synthesize images where the camera is moved left / right, up / down, and front / back with minimal artifacts. 

As shown in \figref{nerf-compare}, we also compare the quality of diffusion-generated images to those generated using the NeRF based approach from SPARTN~\cite{zhou2023nerf}. The moving gripper breaks the static scene assumption made by NeRF. Thus, \cite{zhou2023nerf} masks out the gripper before training the NeRF. 
While this strategy works for the pre-grasp trajectory that \citet{zhou2023nerf} seek to imitate, it fails when the gripper manipulates the scene, as in our tasks. Thus, we investigate different masking schemes: a) no masking, b) masking just the gripper, c) masking a larger region around the gripper; and show visualizations in \figref{nerf-compare}. 

No masking leads to the worst results as expected. Masking just the grabber fixes the grabber, but causes the object being manipulated to blur out. Masking a larger region around the grabber fixes the blurring but also eliminates the object being manipulated. In contrast, images synthesized using our diffusion model move the camera as desired without creating any artifacts. 

\begin{table}[t] 
\centering
\setlength{\tabcolsep}{8pt}
\caption{\textbf{Comparisons between Diffusion-generated Augmented Dataset and Standard Augmentation Schemes.} The table shows median angle error (in radians) between predicted and ground-truth translations. Adding $\Daug$ generated by diffusion models improve performance on top of other augmentation techniques. }
\tablelabel{cmp-aug}
\begin{tabular}{lcc}
\toprule
\bf Methods & \bf BC & \bf \name \\
\midrule
No Aug   & 0.376 & 0.349  \\
w/ Color jitter & 0.381 & 0.350  \\
w/ Flip     &  0.356 &  0.338 \\
w/ Color jitter \& Flip     &  0.355 & \bf 0.328 \\
\bottomrule
\end{tabular}
\end{table}

\begin{figure}[h!]
\centering
\includegraphics[width=1.0\linewidth]{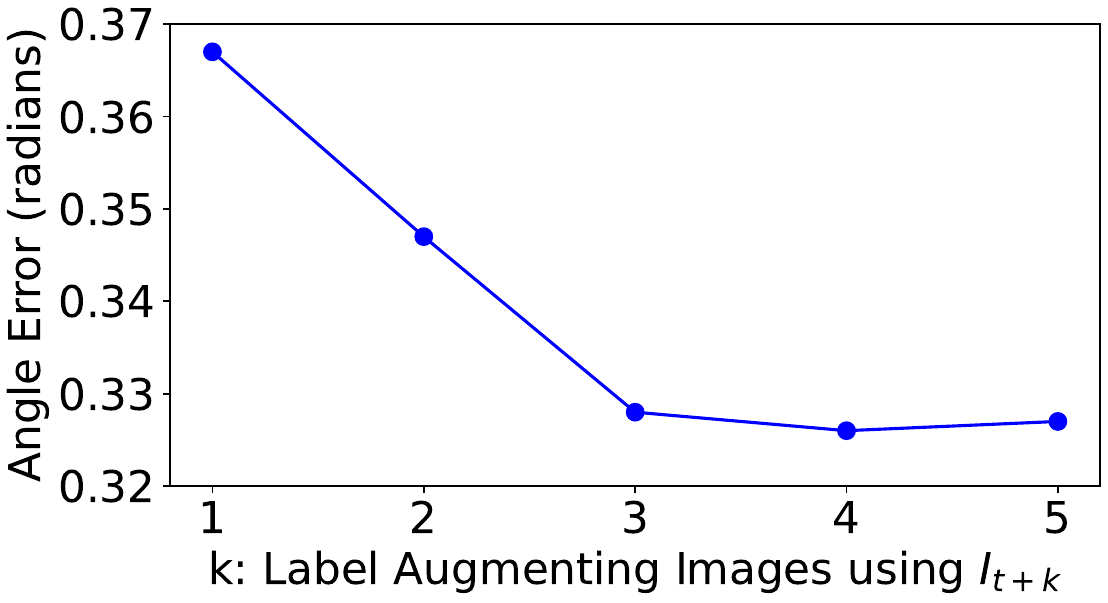}
\caption{{\bf Effect of Using Different Future Frames for Labeling Augmenting Images:} 
We experiment with using different future frame $I_{t+k}$ for labeling the diffusion-generated images. 
Error decreases as $k$ increases and plateaus around $k=3$. Therefore, we use frame $I_{t+3}$ for labeling $\Tilde{I_t}$
}
\figlabel{which-step}
\end{figure} 
\begin{figure*}
\includegraphics[width=1.0\linewidth]{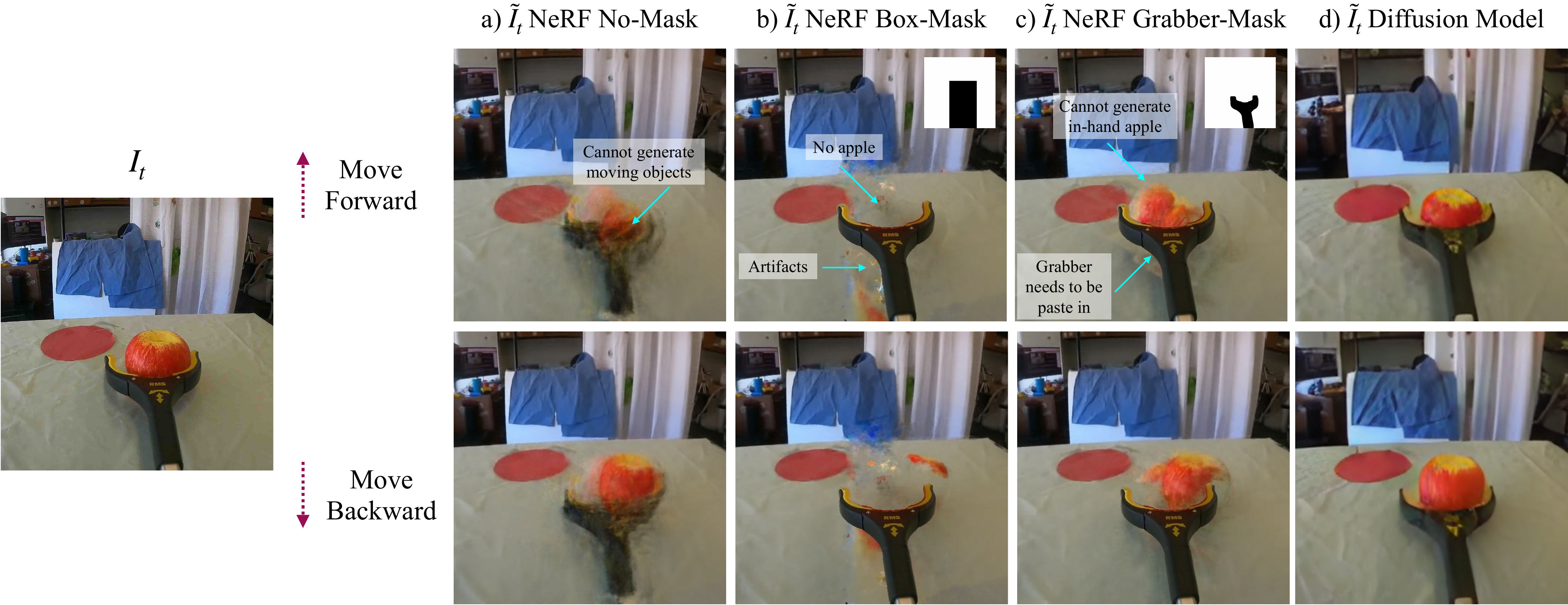}
\caption{\textbf{Diffusion vs NeRF} 
We visualize perturbed samples generated using \name and NeRF with different masking strategies. 
The top row shows images generated for a forward movement relative to $I_t$; the bottom row shows images for a backward movement
{\bf (a) }With no masking, the NeRF reconstructions of the gripper and apple are degenerate, as they violate the static scene assumption in NeRFs. 
{\bf (b)} With a box-mask, the apple is occluded by the mask, and consequently, missing from the reconstruction. 
{\bf (c)} Even with a mask around the end-effector (Grabber-Mask), there are major artifacts in reconstructing the apple.
{\bf (d)} With ours (Diffusion Model), the model faithfully reconstructs the grabber and apple. }
\figlabel{nerf-compare}
\end{figure*}

\subsubsection{\bf Offline Validation}
\seclabel{push-offline}
Out of the 74 expert demonstrations, we use 24, 23, and 27 trajectories for train, validation, and test, respectively.
As done in past work~\cite{young2021visual}, the baseline is vanilla behavior cloning on the expert data.
Our offline validation evaluates the policy's predictions on the test set using the median angle between the predicted and ground truth translations as the error metric.

\paragraph{Comparison with other data augmentation schemes.}
The goal of this experiment is to compare the effectiveness of augmenting with diffusion-generated images to standard techniques, such as color jitter and horizontal flip.
\tableref{cmp-aug} shows that augmenting with out-of-distribution images leads to larger improvement compared to standard techniques. 
Comparing the two columns, we see that adding diffusion-generated images improve performance in all settings.

\begin{table*}
    \centering
    \caption{{\bf Pushing Online Evaluation Results.} 
    (a) \name outperforms BC across all settings.
    \name achieves a 100\% success rate when pushing an apple, greatly exceeding BC's 30\%. It also maintains an 80\% success rate with only 8 demonstrations, whereas BC drops to 20\%.
    (b) As shown in \figref{nerf-compare}, our diffusion model synthesizes higher quality images than NeRFs, especially when scenes undergo deformations. This advantage results in higher task performance: \name achieves a 100\% success rate, while SPARTN~\cite{zhou2023nerf} achieves only 50\%.
    (c) Training the diffusion model with additional play data boosts the task success rate to 100\%, compared to 80\% when using the model trained only on task data.
    }
    \begin{subtable}[t]{0.44\textwidth}
      \centering
      \caption{\bf \name \vs BC}
      \tablelabel{dmd-vs-bc}
      \(\begin{tabular}{lcccc}
        \toprule
        \multirow{2.3}{*}{Method} 
    & \mcc[3]{Apple} & White Cup          \\
    \cmidrule(lr){2-4} \cmidrule(lr){5-5}
    & 24 Demo.   & 16 Demo. & 8 Demo. & 16 Demo. \\
    \midrule
    BC   & 30\%    & 50\%   & 20\%  & 80\%      \\
    DMD    & 100\%    & 90\%   & 80\%  & 90\%    \\
    \bottomrule
    \end{tabular}\)

\end{subtable}
    \begin{subtable}[t]{0.22\textwidth}
      \centering
      \caption{\bf DMD \vs SPARTN~\cite{zhou2023nerf}}
      \tablelabel{dmd-vs-spartn}
      \(\begin{tabular}{lc}
        \toprule
        \multirow{2.3}{*}{Method} 
    & \mcc[1]{Apple}    \\
    \cmidrule(lr){2-2}
& 24 Demo. \\
        \midrule
    SPARTN~\cite{zhou2023nerf}   & 50\%        \\
    DMD    & 100\%         \\
        \bottomrule
        \end{tabular}\)
    \end{subtable}
\begin{subtable}[t]{0.22\textwidth}
      \centering
      \caption{\bf Utility of Play Data}
      \tablelabel{dmd-playutil}
      \(\begin{tabular}{lc}
        \toprule
        \multirow{2.3}{*}{Method} 
    & \multirow{2.3}{*}{Apple}    \\
\vspace{1.7mm}
        &  \\
        \midrule
        DMD Task Only  & 80\%        \\
        DMD Task \& Play   & 100\%         \\
        \bottomrule
        \end{tabular}\)
    \end{subtable}
  \end{table*}
 \paragraph{Overshooting problem, what frame to use?}
\seclabel{overshoot-sec}
In order to mitigate the overshooting problem, we use image $I_{t+k}, k>1$ to label the augmenting image instead of $I_{t+1}$. 
\figref{which-step} shows the effect of this parameter $k$ on policy performance.
We find that labeling with $I_{t+1}$'s pose works worse than labeling with $I_{t+3}$'s pose and beyond. 
As the curve plateaus for $k > 3$ and using larger $k$ leads to the exclusion of the last $k-1$ frames in each trajectory, we use $k = 3$ for our online experiments.

\subsubsection{\bf Online Validation}
\seclabel{online-exp}

\begin{figure*}[t]
\includegraphics[width=\linewidth]{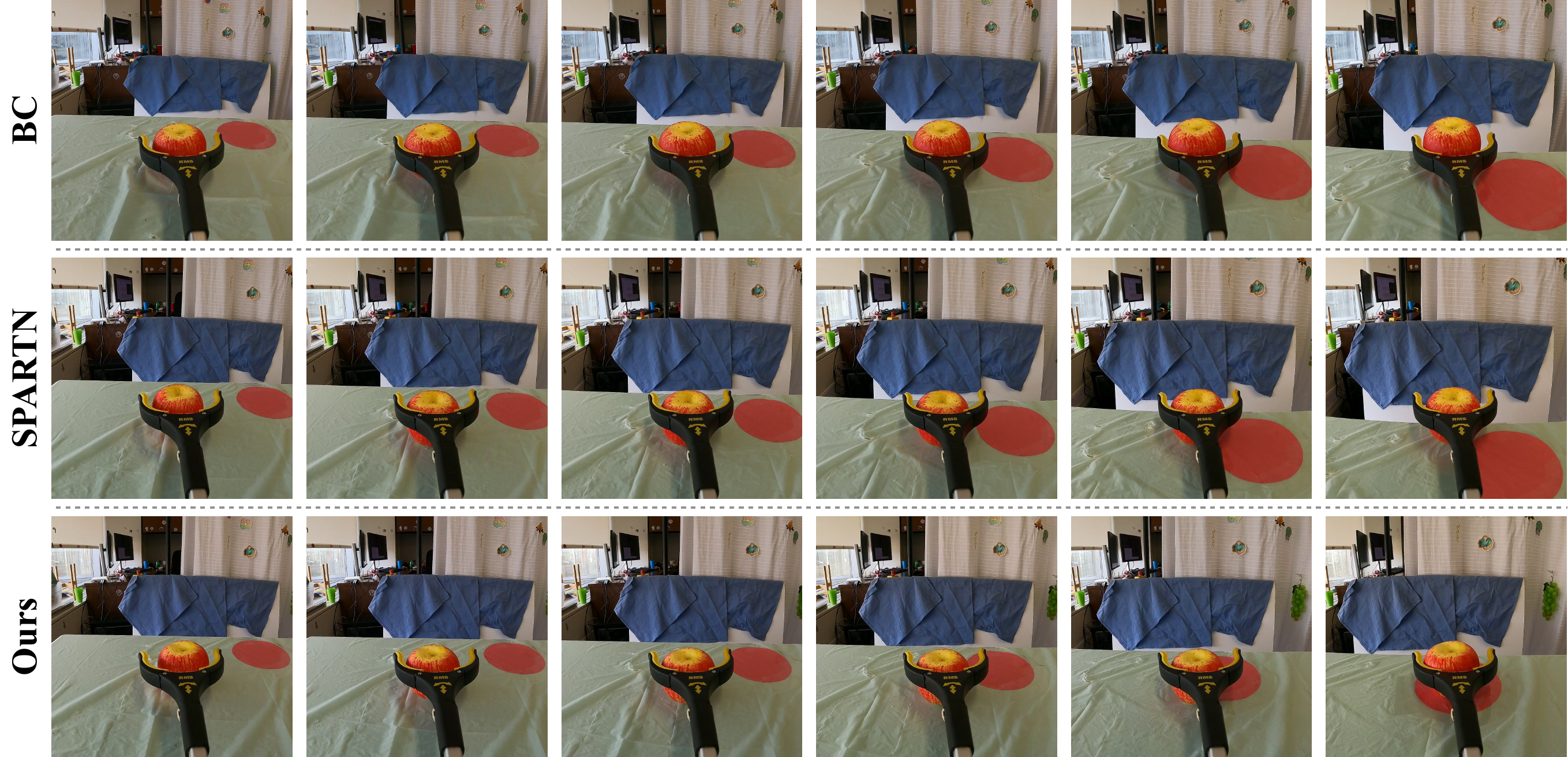} \\
\caption{{\bf Comparison of BC, SPARTN, and \name for Staying on Course.} 
We show the trajectories executed by BC, SPARTN~\cite{zhou2023nerf}, and \name over several steps towards the target.
Unlike BC and SPARTN, which gradually deviate from the intended path, \name maintain its course towards the right and reach the target successfully. See \website for more execution videos.}
\figlabel{recover-offcourse}
\end{figure*}

We conduct 10 trials per method and randomize the apple start location between trials. To prevent human biases, we choose the apple start location before sampling the method to run next. Execution trajectories can be found on the \website.
 
\paragraph{\name \vs Behavior Cloning} 
\seclabel{push_bc}
We take the best model for BC and \name from \tableref{cmp-aug} and evaluate them on the robot. 
We find the vanilla BC model to get a 30\% success rate while \name achieves 100\% (\tableref{dmd-vs-bc}). This significant difference shows that minor error at each step can accumulate and result in task failure.

\paragraph{\name with Few Demonstrations}
\seclabel{few-demo-robot}
Motivated by the 100\% performance of \name with 24 demonstrations, we ask how low can the number of demonstrations be?
We consider 2 low data settings with 8 and 16 demonstrations. For these settings, the diffusion model is trained with only play data.
Success rate for these two models on the robot is shown in \tableref{dmd-vs-bc}. Once again \name outperforms BC and achieves 80\% success even when trained with only 8 demonstrations.

\paragraph{\name \vs SPARTN~\cite{zhou2023nerf}}
\seclabel{spartn}
Following up on the visual comparison of augmenting samples from \secref{vis-gen}, we conduct a head-to-head comparison against SPARTN on the real robot. SPARTN achieves a success rate of 50\% \vs \name succeeds 100\% of the time (\tableref{dmd-vs-spartn}). 
In \figref{recover-offcourse}, we show that while SPARTN diverges from the expert course and pushes the apple off the table, \name brings the apple to the target successfully.  

\paragraph{Utility of Play Data for Training Diffusion Model} 
\seclabel{push_utility_play}
We also experimentally validate what the choice of the dataset necessary for training the view-synthesis diffusion model. We repeat the experiment with 24 demos but vary the diffusion model that generated the augmenting samples. Using the diffusion model trained on just task data lead to an 80\% success rate \vs 100\% success rate with the diffusion model trained on a combination of task and play data (\tableref{dmd-playutil}). Thus, while just training the diffusion model on task data is effective (a nice result in practice), performance can be boosted further by training the diffusion model on play data.

\paragraph{Scaling to Other Objects}
\label{push_cup}
We test \name on another object, a cup. 
The diffusion model is trained with only cup play data.
Task data include 16 demonstrations.  
BC is trained only with task data, and \name is trained additionally with synthesized images.
Similar to apple, we repeat each method 10 times and randomize the cup start location. 
We choose the start location before sampling the method to execute.
The BC baseline achieves a success rate of 80\% vs \name achieves a success rate of 90\% (\tableref{dmd-vs-bc}).

\subsection{\textbf{\textit{Stacking}}}
\seclabel{online-stack}

\subsubsection{\bf Experiment Setup}
Stacking involves reaching a cup, grasping it, placing it on a black box, and opening the grabber.
This task is similar to the stacking task in VIME~\cite{young2021visual} and
requires precision in the height direction to move the cup on the box. 
The action space is the same as in  VIME's publicly available code: relative 3D translations of the end-effector.

We use a grabber and collect 50 demonstrations across three training cups (red dotted box in \figref{old-new-robot-config}(b)).
We obtain cameras poses using ORB-SLAM3~\cite{ORBSLAM3_TRO,chi2024universal}, which recovers the correct metric scale by using the IMU data from GoPro. 

We train the diffusion model on the 50 expert trajectories and don't use any
play data.
Unlike pushing, we do not need to scale by the largest displacement between
adjacent frames because of metric scale from IMU. 
To sample out-of-distribution images, we sample a small translation (with
length between $2cm$ and $4cm$) in a random direction and a small rotation
obtained by sampling yaw, pitch, and roll uniformly from the range
$[-10^{\circ},10^{\circ}]$, $[-10^{\circ},10^{\circ}]$, and
$[-10^{\circ},15^{\circ}]$, respectively.  

The policy architecture is the same as that of pushing and is trained with L1 loss to the normalized ground-truth translation.
We annotate frames at which the grabber open/close and train a gripper policy that is shared across all the methods.

\subsubsection{\bf Online Validation}
We evaluate on the three training cups plus two holdout cups: a larger red cup and a Sprite can (yellow dotted box in \figref{old-new-robot-config}(b)).  
The starting locations of the objects are randomized and do not necessarily match the training data.

\begin{table}[t] 
\centering
\caption{\textbf{Stacking Online Evaluation Results.} In the task of stacking a cup on a
box, \name achieves over 90\% success rate for all training cups and over
80\% for the two cups unseen cups and outperforms BC. 
The diffusion model is only trained on task data/expert demonstrations.
See all execution videos
on \website.
}
\tablelabel{stack}
\begin{tabular}{lccccc}
\toprule
& \multicolumn{3}{c}{\bf Seen Cups} & \multicolumn{2}{c}{\bf Unseen Cups} \\
\cmidrule(lr){2-4} \cmidrule(lr){5-6}
\bf Method & \bf  Blue & \bf  Red-1 & \bf  White & \bf Red-2 & \bf Sprite \\
\midrule
BC & 10\%  & 10\% & 90\% & 40\% & 50\% \\
\name (Task Only) & \textbf{100\%}  & 
\textbf{90\%} & 
\textbf{100\%} & 
\textbf{80\%} & 
\textbf{90\%} \\
\bottomrule
\end{tabular}
\end{table}

As shown in \tableref{stack}, \name performs better than BC across both seen and unseen instances.
BC often fails to lift the tall cups above the box and pushes the box forward continuously. 
See videos on \website for failure modes.
This behavior might explain the higher success rate of the white cup, which is shorter and thus easier to lift above the box.
By training on synthesized data, \name learns to lower the grabber if approaching the cups too high, and it grasps closer to the cup bottom so that moving above the box becomes easier.
It succeeds more than BC across all cups.

\subsection{\textbf{\textit{Pouring}}}
\seclabel{online-pour}

\subsubsection{\bf Experiment Setup}
As shown in \figref{old-new-robot-config}(c), this task requires the robot to reach a blue cup filled with 57 grams of coffee beans, grasp it, and pour all the beans into a red cup. 
Success is determined by the weight of the coffee beans being the same at the end of each trial.
Action space is full 6-DoF.

Task data contain 49 demonstrations, and play data contain 16 trajectories. 
Diffusion model is trained on both types of data.
Except for an additional MLP-head to predict rotation, the policy architecture is the same as that of stacking; the model is trained with L1 loss to the ground-truth translation and rotation. 

During execution, the robot executes the translation and rotation predicted by the model.
Locations of the two cups are randomized, but the blue cup is always on the same side of the red cup.

\begin{table}[t] 
    \centering
\setlength{\tabcolsep}{8pt}
\caption{\textbf{Pouring and Hanging-Shirt Online Evaluation Results.} \name outperforms BC
across both tasks. 
See all execution videos on \website. $^{\ast}$: the two BC numbers are different because they are from two
different pairwise randomized A/B tests: 1) BC \vs DMD (Task \& Play) and 2) BC \vs
DMD (Task Only), and experimental conditions (\eg lighting) may have been different
between when these two tests were conducted.}
\tablelabel{pour-hanger}
\begin{tabular}{lccc}
\toprule
& \multicolumn{2}{c}{\bf Pouring} & \bf Hanging \\
\cmidrule(lr){2-3} \cmidrule(lr){4-4}
\bf Method & \bf  Task \& Play & \bf Task Only & \bf  Task \& Play \\
\midrule
BC & $0\%^{\ast}$  & $30\%^{\ast}$ & 20\% \\
\name & \textbf{80\%}  & \textbf{70\%} & 
\textbf{90\%} \\
\bottomrule
\end{tabular}
\end{table}

\subsubsection{\bf Online Validation}
\seclabel{pour-online-exp}
\name succeeds 80\% of the time while BC fails completely.
A common failure case for BC is that as the robot rotates the cup with coffee beans, it does not move the cup closer to the receiving cup; the blue cup then drifts further and further from the one on the table. 
In contrast, even in the worst trial, \name can pour 
53 grams (\ie more than 90\%) of the coffee beans. 
See execution videos on \website for clear differences between BC and \name.

We also validate the use of play data for learning an effective diffusion model. 
We finetune the diffusion model with only task data and conduct the same
experiment as the one in the last paragraph.
As shown in \tableref{pour-hanger}, without training the diffusion model with play data, the downstream policy maintains a 70\% success rate, higher than BC's 30\%. 
This shows that, training on task data alone also enables the diffusion model
to synthesize good out-of-distribution images that improve task performance.
Note that the two BC numbers are different because they are from two different
pairwise randomized A/B tests: 1) BC \vs DMD (Task \& Play) and 2) BC \vs DMD (Task
Only), and experimental conditions (\eg lighting) may have been different
between when these tests were conducted.

\subsection{\textbf{\textit{Hanging a Shirt}}}
\seclabel{online-hanger}

\subsubsection{\bf Experiment Setup}
To hang the shirt, the robot reaches the hanger and slides the shirt across the table to the edge; it then grasps the part of the hanger sticking out of the table and hangs the shirt on the clothing rack.
The setup is shown in \figref{old-new-robot-config}(d).
Action space is full 6-DoF.

Task data contains 55 demonstrations, and play data contains 24 trajectories. 
The diffusion model is trained on both types of data.
We divide the hanging task into two sub-tasks: grabbing the shirt and putting
it on the rack. A separate policy is learned for each task.
We use same policy architecture and loss function as those used in pouring.

During execution, the robot starts with the grabbing policy. 
When the network predicts to close the grabber, the grabbing policy terminates,
and the robot switches to the putting-on-rack policy. The entire process
terminates when the top of the hanger makes contact with the rack pole,
signaled by when the end-effector senses a force greater than $4N$.  
We add some clay padding to the grabber tips because the hanger is thin, and the motor is not strong enough to close the grabber completely.  

\subsubsection{\bf Online Validation}
As shown in \tableref{pour-hanger}, \name achieves a 90\% success rate while BC
only succeed in 20\% cases. 
BC often fails because the robot raises the shirt too high, and the shoulder
part of the hanger hits the pole before the hook part reaches the pole.
On the other hand, \name learns to bring the shirt down onto the rack.
 
\subsection{\textbf{\textit{In-the-Wild Cup Arrangement}}}
\seclabel{cup-arrangement}

\begin{figure}[t]
\includegraphics[width=\linewidth]{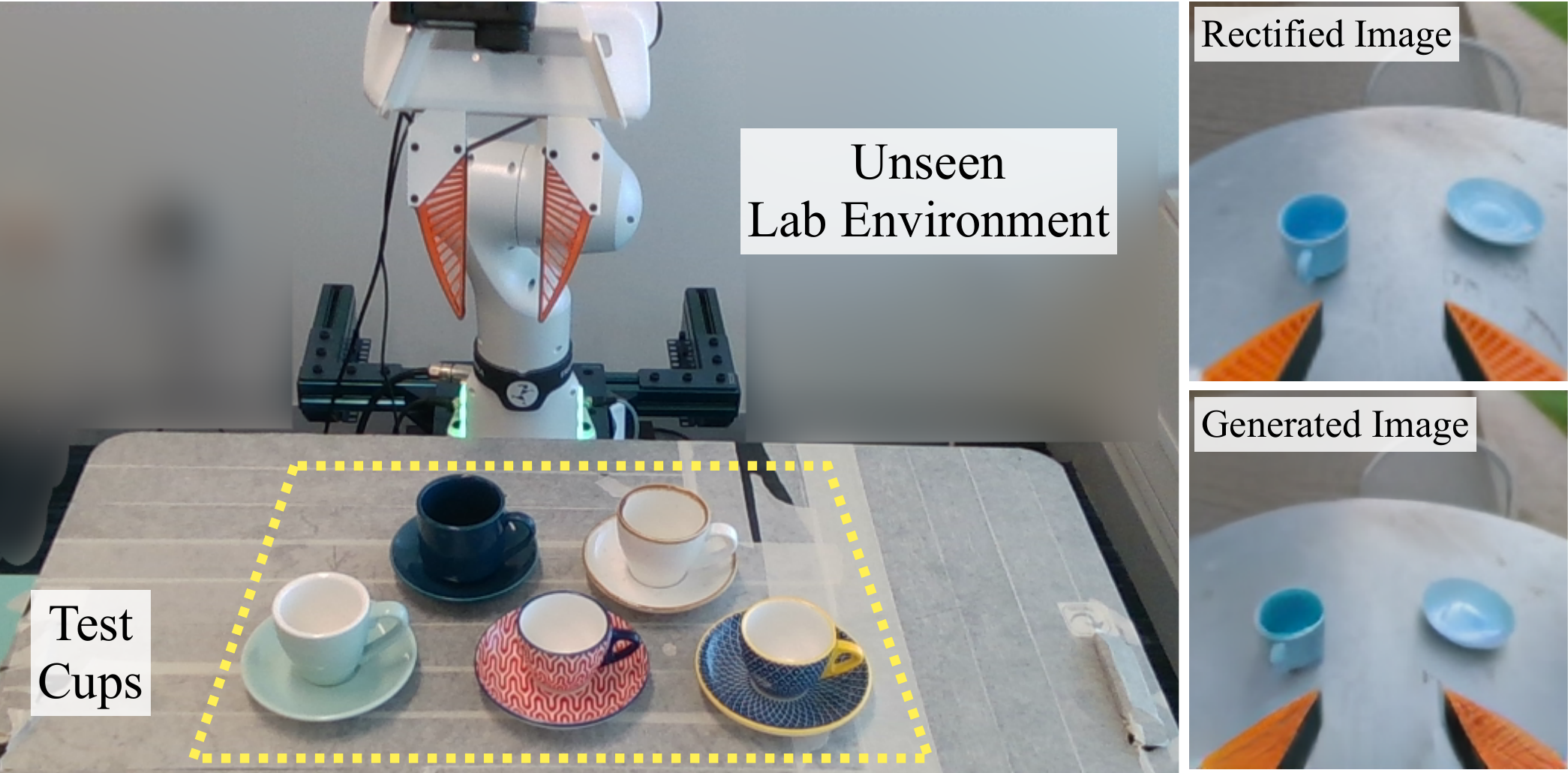}
\caption{{\bf In-the-Wild Cup Arrangement Setup:} 
We evaluate \name on the in-the-wild cup arrangement task from
UMI~\cite{chi2024universal}. The training set for this task contains 1447
demonstrations across 30 locations and 18 cups. Because our pre-trained
diffusion model operates on non-fish eye images, we work with center-cropped
and rectified images  from the Fisheye lens. Baseline and baseline+\name
operate on these center-cropped rectified images. Evaluation is done on 5 novel
cups in a novel environment (our lab).}
\figlabel{umi-cup}
\end{figure}

\begin{table}[t] 
\centering
\setlength{\tabcolsep}{4pt}
\caption{\textbf{In-the-Wild Cup Arrangement Online Evaluation Results.} 
\name generalizes better to novel cups in novel environment than vanilla
Diffusion Policy~\cite{chi2023diffusionpolicy}, achieving a 64\% overall
success rate compared to Diffusion Policy's 16\%. It succeeds in same or more
trials across all 5 cups.  See all execution videos on \website.
}
\tablelabel{cup}
\begin{tabular}{lcccccc}
\toprule
\bf Method & \bf  Navy & \bf  White & \bf  Mint & \bf Red & \bf Blue & \bf  Overall   \\\midrule
Diffusion Policy~\cite{chi2023diffusionpolicy} 
& 1/5  & 1/5 & 1/5 & 0/5 & 1/5 & 4/25 (16\%) \\
\name (Task Only) 
& \bf 3/5  & \bf 3/5 & \bf 3/5 & \bf 3/5 & \bf 4/5 & \bf 16/25 (64\%) \\
\bottomrule
\end{tabular}
\end{table}
 
The goal of this experiment is to understand a) can \name improve performance
in the presence of a large number of diverse demonstrations, b) can \name
improve generalization of policies to novel objects in novel environments, and
c) if \name is compatible with policies trained with
diffusion~\cite{chi2023diffusionpolicy}.

\subsubsection{\bf Experiment Setup}
We leverage a diverse in-the-wild dataset from the recent Universal
Manipulation Interface (UMI) paper~\cite{chi2024universal}. 
It introduces a hand-held data collection device and an imitation learning
framework utilizing Diffusion Policy~\cite{chi2023diffusionpolicy}.
The device captures image observations using a wrist-mounted GoPro camera.

We adopt the same task definition as the in-the-wild generalization experiment
in UMI: placing a cup on a saucer with its handle facing the left side of the
robot. UMI~\cite{chi2024universal} collected 1447 demonstrations across 30
locations and 18 training cups. We use their publicly available demonstration
data and conduct evaluation in our lab (\ie novel location) with and without
\name.

Our evaluation setup is identical to that used in UMI, with one difference. UMI
uses wide field-of-view images obtained by putting an additional Fisheye
lens on the GoPro. Our diffusion model denoises in the latent space of a VQ-GAN
autoencoder~\cite{9578911} that was pre-trained on the RealEstate10K
dataset~\cite{10.1145/3197517.3201323}, which contains non fish-eye images.
Therefore, our diffusion model is constrained to only work well on non-fish eye
images. Thus, we conduct comparisons where both the baseline and baseline+\name
operate on center-cropped rectified images shown in \figref{umi-cup}. We
retrain the baseline diffusion policy from UMI to consume these center-cropped
rectified images.
Other than this change, the task policy uses UMI's implementation of Diffusion
Policy~\cite{chi2023diffusionpolicy} and denoises 6DoF actions.

We test on 5 held-out cups shown in \figref{umi-cup}. These cups are not used
for expert demonstrations in dataset from \cite{chi2024universal}. For each
  cup, we test 5 different start configurations. We follow the experiment
  protocol outlined in \cite{chi2024universal}: we use pixel masks to make sure
  that the starting locations of the cups and saucers are the same across the
  two methods.

\subsubsection{\bf Online Validation}
\name improves upon the vanilla Diffusion Policy~\cite{chi2023diffusionpolicy}
across all novel cups, achieving an overall success rate of 64\%, compared to
Diffusion Policy's 16\% success rate (\tableref{cup}).  This improvement shows
that \name is effective even in the presence of large number (1447) of diverse
demonstrations and also leads to improved performance when testing on novel
objects in novel environments. Furthermore, \name-created data also improves
policies trained via diffusion. 
One common failure case for the baseline is that the policy doesn't reorient
the cup correctly, because it fails to make contact with the cup or its handle.
\name, on the other hand, successfully pushes the handle to the left side more
often and grasps the cup without tipping it over. 
  
\section{Discussion}
\seclabel{discussion}
We developed \name, a data creation framework that improves the sample
efficiency of eye-in-hand imitation learning.  
\name uses a diffusion model to synthesize out-of-distribution 
views and assigns them corrective labels. Additional use of these
out-of-distribution views for policy training leads to more performant
policies. \name leads to large benefits over traditional behavior cloning across
4 diverse tasks: pushing, stacking, pouring, and hanging a shirt. \name attains
80\% success rate from as few as 8 demonstrations in pushing and reaches an
average of 92\% success rate across 5 different cups in stacking. When pouring,
it transfers coffee beans successfully 80\% of the time; finally, it can slide
a shirt off the table and put it on the rack with 90\% success rate. Together
these tasks test DMD on different aspects: 3DoF action space (pushing,
stacking), 6DoF action spaces (pouring, hanging a shirt), generalization to new
objects (stacking), and precision in reaching goal locations (pouring, hanging a shirt).

Our experiments reveal the benefits of the various parts of \name over past
approaches.  The use of diffusion models to synthesize images allows \name to generate augmenting images for manipulation tasks involving a non-static scene.  Prior work~\cite{zhou2023nerf} used NeRFs instead of diffusion models,
restricting generation to samples from only the pre-grasp part of the
demonstration.
We showcase the advantages of diffusion models over NeRFs through qualitative
results in \figref{nerf-compare} and real robot experiments in \secref{spartn}.

We also tested which type of data: task data, play data, or a combination of
both, should be used for training the diffusion model.  We found all versions
to outperform behavior cloning (\secref{push_utility_play}, \tableref{stack},
and \tableref{pour-hanger}).  This suggests two practically useful results.
First, \name can be directly used to improve a task policy,
\ie the task data itself can be used to finetune a novel-view synthesis
diffusion model, augmenting the task dataset to improve performance.
Second, we can utilize play data, which may be simpler to collect than task data.

\name also has several limitations.
\name assumes that the generated states are recoverable, \ie there exists an action that can return to the expert distribution. 
This is not always true, for example, when the robot rotates the
cup at the wrong location and the coffee beans spill. 
However, \name can still be useful if it can bring the system back to
in-distribution states before the system spirals into irrecoverable state.

Possible future directions include: learning diffusion models that can predict
highly discontinuous object dynamics, generating out-of-distribution states of
non-visual modalities (\eg forces), and 
finding ways to assign action labels in situations where they cannot be derived from relative camera transformation.
Code, data and models are publicly available on the \website.

\section*{Acknowledgments}
We give special thanks to Kevin Zhang for 3D printing the attachment for us. This material is based upon work supported by the USDA-NIFA AIFARMS National AI Institute (USDA \#2020-67021-32799) and NSF (IIS-2007035).

\bibliographystyle{plainnat}
\bibliography{biblioLong, references, references_sg}

\begin{thebibliography}{88}
\providecommand{\natexlab}[1]{#1}
\providecommand{\url}[1]{\texttt{#1}}
\expandafter\ifx\csname urlstyle\endcsname\relax
  \providecommand{\doi}[1]{doi: #1}\else
  \providecommand{\doi}{doi: \begingroup \urlstyle{rm}\Url}\fi

\bibitem[Ada et~al.(2024)Ada, Oztop, and Ugur]{ada2024diffusion}
Suzan~Ece Ada, Erhan Oztop, and Emre Ugur.
\newblock Diffusion policies for out-of-distribution generalization in offline
  reinforcement learning.
\newblock \emph{IEEE Robotics and Automation Letters}, 2024.

\bibitem[Adamkiewicz et~al.(2022)Adamkiewicz, Chen, Caccavale, Gardner,
  Culbertson, Bohg, and Schwager]{adamkiewicz2022vision}
Michal Adamkiewicz, Timothy Chen, Adam Caccavale, Rachel Gardner, Preston
  Culbertson, Jeannette Bohg, and Mac Schwager.
\newblock Vision-only robot navigation in a neural radiance world.
\newblock \emph{IEEE Robotics and Automation Letters}, 7\penalty0 (2):\penalty0
  4606--4613, 2022.

\bibitem[Ajay et~al.(2022)Ajay, Du, Gupta, Tenenbaum, Jaakkola, and
  Agrawal]{ajay2022conditional}
Anurag Ajay, Yilun Du, Abhi Gupta, Joshua Tenenbaum, Tommi Jaakkola, and Pulkit
  Agrawal.
\newblock Is conditional generative modeling all you need for decision-making?
\newblock \emph{arXiv preprint arXiv:2211.15657}, 2022.

\bibitem[Argall et~al.(2009)Argall, Chernova, Veloso, and
  Browning]{argall2009survey}
Brenna~D Argall, Sonia Chernova, Manuela Veloso, and Brett Browning.
\newblock A survey of robot learning from demonstration.
\newblock \emph{Robotics and autonomous systems}, 57\penalty0 (5):\penalty0
  469--483, 2009.

\bibitem[Balaji et~al.(2022)Balaji, Nah, Huang, Vahdat, Song, Kreis, Aittala,
  Aila, Laine, Catanzaro, et~al.]{balaji2022ediffi}
Yogesh Balaji, Seungjun Nah, Xun Huang, Arash Vahdat, Jiaming Song, Karsten
  Kreis, Miika Aittala, Timo Aila, Samuli Laine, Bryan Catanzaro, et~al.
\newblock ediffi: Text-to-image diffusion models with an ensemble of expert
  denoisers.
\newblock \emph{arXiv preprint arXiv:2211.01324}, 2022.

\bibitem[Black et~al.(2023)Black, Nakamoto, Atreya, Walke, Finn, Kumar, and
  Levine]{black2023zero}
Kevin Black, Mitsuhiko Nakamoto, Pranav Atreya, Homer Walke, Chelsea Finn,
  Aviral Kumar, and Sergey Levine.
\newblock Zero-shot robotic manipulation with pretrained image-editing
  diffusion models.
\newblock \emph{arXiv preprint arXiv:2310.10639}, 2023.

\bibitem[Blukis et~al.(2022)Blukis, Lee, Tremblay, Wen, Kweon, Yoon, Fox, and
  Birchfield]{blukis2022neural}
Valts Blukis, Taeyeop Lee, Jonathan Tremblay, Bowen Wen, In~So Kweon, Kuk-Jin
  Yoon, Dieter Fox, and Stan Birchfield.
\newblock Neural fields for robotic object manipulation from a single image.
\newblock \emph{arXiv preprint arXiv:2210.12126}, 2022.

\bibitem[Campos et~al.(2021)Campos, Elvira, G\'omez, Montiel, and
  Tard\'os]{ORBSLAM3_TRO}
Carlos Campos, Richard Elvira, Juan~J. G\'omez, Jos\'e M.~M. Montiel, and
  Juan~D. Tard\'os.
\newblock {ORB-SLAM3}: An accurate open-source library for visual,
  visual-inertial and multi-map {SLAM}.
\newblock \emph{IEEE Transactions on Robotics}, 37\penalty0 (6):\penalty0
  1874--1890, 2021.

\bibitem[Cao and Johnson(2023)]{cao2023hexplane}
Ang Cao and Justin Johnson.
\newblock Hexplane: A fast representation for dynamic scenes.
\newblock In \emph{Proceedings of the IEEE Conference on Computer Vision and
  Pattern Recognition (CVPR)}, pages 130--141, 2023.

\bibitem[Chang et~al.(2023)Chang, Prakash, and Gupta]{chang2023look}
Matthew Chang, Aditya Prakash, and Saurabh Gupta.
\newblock Look ma, no hands! agent-environment factorization of egocentric
  videos.
\newblock In \emph{Advances in Neural Information Processing Systems
  (NeurIPS)}, 2023.

\bibitem[Chen et~al.(2024)Chen, Deng, Kawaguchi, Gulcehre, and
  Ahn]{chen2024simple}
Chang Chen, Fei Deng, Kenji Kawaguchi, Caglar Gulcehre, and Sungjin Ahn.
\newblock Simple hierarchical planning with diffusion.
\newblock \emph{arXiv preprint arXiv:2401.02644}, 2024.

\bibitem[Chen et~al.(2023{\natexlab{a}})Chen, Bahl, and
  Pathak]{chen2023playfusion}
Lili Chen, Shikhar Bahl, and Deepak Pathak.
\newblock Playfusion: Skill acquisition via diffusion from language-annotated
  play.
\newblock In \emph{Conference on Robot Learning}, pages 2012--2029. PMLR,
  2023{\natexlab{a}}.

\bibitem[Chen et~al.(2023{\natexlab{b}})Chen, Kiami, Gupta, and
  Kumar]{chen2023genaug}
Zoey Chen, Sho Kiami, Abhishek Gupta, and Vikash Kumar.
\newblock Genaug: Retargeting behaviors to unseen situations via generative
  augmentation.
\newblock \emph{arXiv preprint arXiv:2302.06671}, 2023{\natexlab{b}}.

\bibitem[Chi et~al.(2023)Chi, Feng, Du, Xu, Cousineau, Burchfiel, and
  Song]{chi2023diffusionpolicy}
Cheng Chi, Siyuan Feng, Yilun Du, Zhenjia Xu, Eric Cousineau, Benjamin
  Burchfiel, and Shuran Song.
\newblock Diffusion policy: Visuomotor policy learning via action diffusion.
\newblock In \emph{Robotics: Science and Systems (RSS)}, 2023.

\bibitem[Chi et~al.(2024)Chi, Xu, Pan, Cousineau, Burchfiel, Feng, Tedrake, and
  Song]{chi2024universal}
Cheng Chi, Zhenjia Xu, Chuer Pan, Eric Cousineau, Benjamin Burchfiel, Siyuan
  Feng, Russ Tedrake, and Shuran Song.
\newblock Universal manipulation interface: In-the-wild robot teaching without
  in-the-wild robots.
\newblock \emph{arXiv preprint arXiv:2402.10329}, 2024.

\bibitem[Esser et~al.(2021{\natexlab{a}})Esser, Rombach, and Ommer]{9578911}
Patrick Esser, Robin Rombach, and Björn Ommer.
\newblock Taming transformers for high-resolution image synthesis.
\newblock In \emph{Proceedings of the IEEE Conference on Computer Vision and
  Pattern Recognition (CVPR)}, pages 12868--12878, 2021{\natexlab{a}}.

\bibitem[Esser et~al.(2021{\natexlab{b}})Esser, Rombach, and
  Ommer]{esser2020taming}
Patrick Esser, Robin Rombach, and Björn Ommer.
\newblock Taming transformers for high-resolution image synthesis.
\newblock In \emph{Proceedings of the IEEE Conference on Computer Vision and
  Pattern Recognition (CVPR)}, 2021{\natexlab{b}}.

\bibitem[Florence et~al.(2019)Florence, Manuelli, and
  Tedrake]{florence2019self}
Peter Florence, Lucas Manuelli, and Russ Tedrake.
\newblock Self-supervised correspondence in visuomotor policy learning.
\newblock \emph{IEEE Robotics and Automation Letters}, 5\penalty0 (2):\penalty0
  492--499, 2019.

\bibitem[Giusti et~al.(2016)Giusti, Guzzi, Cireşan, He, Rodríguez, Fontana,
  Faessler, Forster, Schmidhuber, Caro, Scaramuzza, and Gambardella]{7358076}
Alessandro Giusti, Jérôme Guzzi, Dan~C. Cireşan, Fang-Lin He, Juan~P.
  Rodríguez, Flavio Fontana, Matthias Faessler, Christian Forster, Jürgen
  Schmidhuber, Gianni~Di Caro, Davide Scaramuzza, and Luca~M. Gambardella.
\newblock A machine learning approach to visual perception of forest trails for
  mobile robots.
\newblock \emph{IEEE Robotics and Automation Letters}, 1\penalty0 (2):\penalty0
  661--667, 2016.
\newblock \doi{10.1109/LRA.2015.2509024}.

\bibitem[Goodfellow et~al.(2014)Goodfellow, Pouget-Abadie, Mirza, Xu,
  Warde-Farley, Ozair, Courville, and Bengio]{goodfellow2014generative}
Ian Goodfellow, Jean Pouget-Abadie, Mehdi Mirza, Bing Xu, David Warde-Farley,
  Sherjil Ozair, Aaron Courville, and Yoshua Bengio.
\newblock Generative adversarial nets.
\newblock In \emph{Advances in Neural Information Processing Systems
  (NeurIPS)}, volume~27, 2014.

\bibitem[Gu et~al.(2023)Gu, Trevithick, Lin, Susskind, Theobalt, Liu, and
  Ramamoorthi]{gu2023nerfdiff}
Jiatao Gu, Alex Trevithick, Kai-En Lin, Joshua~M Susskind, Christian Theobalt,
  Lingjie Liu, and Ravi Ramamoorthi.
\newblock Nerfdiff: Single-image view synthesis with nerf-guided distillation
  from 3d-aware diffusion.
\newblock In \emph{International Conference on Machine Learning}, pages
  11808--11826. PMLR, 2023.

\bibitem[Gu et~al.(2022)Gu, Chen, Bao, Wen, Zhang, Chen, Yuan, and
  Guo]{gu2022vector}
Shuyang Gu, Dong Chen, Jianmin Bao, Fang Wen, Bo~Zhang, Dongdong Chen, Lu~Yuan,
  and Baining Guo.
\newblock Vector quantized diffusion model for text-to-image synthesis.
\newblock In \emph{Proceedings of the IEEE/CVF Conference on Computer Vision
  and Pattern Recognition}, pages 10696--10706, 2022.

\bibitem[Hartley and Zisserman(2003)]{hartley2003multiple}
Richard Hartley and Andrew Zisserman.
\newblock \emph{Multiple view geometry in computer vision}.
\newblock Cambridge university press, 2003.

\bibitem[He et~al.(2016)He, Zhang, Ren, and Sun]{he2015deep}
Kaiming He, Xiangyu Zhang, Shaoqing Ren, and Jian Sun.
\newblock Deep residual learning for image recognition.
\newblock In \emph{Proceedings of the IEEE Conference on Computer Vision and
  Pattern Recognition (CVPR)}, pages 770--778, 2016.

\bibitem[Hegde et~al.(2023)Hegde, Batra, Zentner, and
  Sukhatme]{hegde2023generating}
Shashank Hegde, Sumeet Batra, KR~Zentner, and Gaurav~S Sukhatme.
\newblock Generating behaviorally diverse policies with latent diffusion
  models.
\newblock \emph{arXiv preprint arXiv:2305.18738}, 2023.

\bibitem[Ho et~al.(2020)Ho, Jain, and Abbeel]{ho2020denoising}
Jonathan Ho, Ajay Jain, and Pieter Abbeel.
\newblock Denoising diffusion probabilistic models.
\newblock \emph{Advances in Neural Information Processing Systems},
  33:\penalty0 6840--6851, 2020.

\bibitem[Hussein et~al.(2017)Hussein, Gaber, Elyan, and
  Jayne]{hussein2017imitation}
Ahmed Hussein, Mohamed~Medhat Gaber, Eyad Elyan, and Chrisina Jayne.
\newblock Imitation learning: A survey of learning methods.
\newblock \emph{ACM Computing Surveys (CSUR)}, 50\penalty0 (2):\penalty0 1--35,
  2017.

\bibitem[Isola et~al.(2017)Isola, Zhu, Zhou, and Efros]{isola2017image}
Phillip Isola, Jun-Yan Zhu, Tinghui Zhou, and Alexei~A Efros.
\newblock Image-to-image translation with conditional adversarial networks.
\newblock In \emph{Proceedings of the IEEE Conference on Computer Vision and
  Pattern Recognition (CVPR)}, pages 1125--1134, 2017.

\bibitem[Jang et~al.(2022)Jang, Irpan, Khansari, Kappler, Ebert, Lynch, Levine,
  and Finn]{jang2022bc}
Eric Jang, Alex Irpan, Mohi Khansari, Daniel Kappler, Frederik Ebert, Corey
  Lynch, Sergey Levine, and Chelsea Finn.
\newblock {BC-Z}: Zero-shot task generalization with robotic imitation
  learning.
\newblock In \emph{Proceedings of the Conference on Robot Learning (CoRL)},
  pages 991--1002. PMLR, 2022.

\bibitem[Janner et~al.(2022)Janner, Du, Tenenbaum, and
  Levine]{janner2022planning}
Michael Janner, Yilun Du, Joshua~B Tenenbaum, and Sergey Levine.
\newblock Planning with diffusion for flexible behavior synthesis.
\newblock \emph{arXiv preprint arXiv:2205.09991}, 2022.

\bibitem[Kapelyukh et~al.(2023)Kapelyukh, Vosylius, and
  Johns]{kapelyukh2023dall}
Ivan Kapelyukh, Vitalis Vosylius, and Edward Johns.
\newblock Dall-e-bot: Introducing web-scale diffusion models to robotics.
\newblock \emph{IEEE Robotics and Automation Letters}, 2023.

\bibitem[Ke et~al.(2021)Ke, Wang, Bhattacharjee, Boots, and
  Srinivasa]{ke2021grasping}
Liyiming Ke, Jingqiang Wang, Tapomayukh Bhattacharjee, Byron Boots, and
  Siddhartha Srinivasa.
\newblock Grasping with chopsticks: Combating covariate shift in model-free
  imitation learning for fine manipulation.
\newblock In \emph{Proceedings of the IEEE International Conference on Robotics
  and Automation (ICRA)}, pages 6185--6191. IEEE, 2021.

\bibitem[Krizhevsky et~al.(2012)Krizhevsky, Sutskever, and
  Hinton]{10.5555/2999134.2999257}
Alex Krizhevsky, Ilya Sutskever, and Geoffrey~E. Hinton.
\newblock Imagenet classification with deep convolutional neural networks.
\newblock In \emph{Advances in Neural Information Processing Systems
  (NeurIPS)}, page 1097–1105, 2012.

\bibitem[Kwak et~al.(2023)Kwak, Dong, Jin, Ko, Mahajan, and Yi]{kwak2023vivid}
Jeong-gi Kwak, Erqun Dong, Yuhe Jin, Hanseok Ko, Shweta Mahajan, and Kwang~Moo
  Yi.
\newblock Vivid-1-to-3: Novel view synthesis with video diffusion models.
\newblock \emph{arXiv preprint arXiv:2312.01305}, 2023.

\bibitem[Laskey et~al.(2016)Laskey, Staszak, Hsieh, Mahler, Pokorny, Dragan,
  and Goldberg]{laskey2016shiv}
Michael Laskey, Sam Staszak, Wesley Yu-Shu Hsieh, Jeffrey Mahler, Florian~T
  Pokorny, Anca~D Dragan, and Ken Goldberg.
\newblock Shiv: Reducing supervisor burden in dagger using support vectors for
  efficient learning from demonstrations in high dimensional state spaces.
\newblock In \emph{Proceedings of the IEEE International Conference on Robotics
  and Automation (ICRA)}, pages 462--469. IEEE, 2016.

\bibitem[Laskey et~al.(2017)Laskey, Lee, Fox, Dragan, and
  Goldberg]{laskey2017dart}
Michael Laskey, Jonathan Lee, Roy Fox, Anca Dragan, and Ken Goldberg.
\newblock Dart: Noise injection for robust imitation learning.
\newblock In \emph{Proceedings of the Conference on Robot Learning (CoRL)},
  pages 143--156. PMLR, 2017.

\bibitem[Li et~al.(2023{\natexlab{a}})Li, Wang, Jin, and
  Zha]{li2023hierarchical}
Wenhao Li, Xiangfeng Wang, Bo~Jin, and Hongyuan Zha.
\newblock Hierarchical diffusion for offline decision making.
\newblock In \emph{Proceedings of the International Conference on Machine
  Learning (ICML)}, pages 20035--20064. PMLR, 2023{\natexlab{a}}.

\bibitem[Li et~al.(2023{\natexlab{b}})Li, Belagali, Shang, and
  Ryoo]{li2023crossway}
Xiang Li, Varun Belagali, Jinghuan Shang, and Michael~S Ryoo.
\newblock Crossway diffusion: Improving diffusion-based visuomotor policy via
  self-supervised learning.
\newblock \emph{arXiv preprint arXiv:2307.01849}, 2023{\natexlab{b}}.

\bibitem[Li et~al.(2022)Li, Li, Sitzmann, Agrawal, and Torralba]{li20223d}
Yunzhu Li, Shuang Li, Vincent Sitzmann, Pulkit Agrawal, and Antonio Torralba.
\newblock 3d neural scene representations for visuomotor control.
\newblock In \emph{Conference on Robot Learning}, pages 112--123. PMLR, 2022.

\bibitem[Liang et~al.(2023{\natexlab{a}})Liang, Mu, Ding, Ni, Tomizuka, and
  Luo]{liang2023adaptdiffuser}
Zhixuan Liang, Yao Mu, Mingyu Ding, Fei Ni, Masayoshi Tomizuka, and Ping Luo.
\newblock Adaptdiffuser: Diffusion models as adaptive self-evolving planners.
\newblock \emph{arXiv preprint arXiv:2302.01877}, 2023{\natexlab{a}}.

\bibitem[Liang et~al.(2023{\natexlab{b}})Liang, Mu, Ma, Tomizuka, Ding, and
  Luo]{liang2023skilldiffuser}
Zhixuan Liang, Yao Mu, Hengbo Ma, Masayoshi Tomizuka, Mingyu Ding, and Ping
  Luo.
\newblock Skilldiffuser: Interpretable hierarchical planning via skill
  abstractions in diffusion-based task execution.
\newblock \emph{arXiv preprint arXiv:2312.11598}, 2023{\natexlab{b}}.

\bibitem[Liu et~al.(2023)Liu, Wu, Van~Hoorick, Tokmakov, Zakharov, and
  Vondrick]{liu2023zero}
Ruoshi Liu, Rundi Wu, Basile Van~Hoorick, Pavel Tokmakov, Sergey Zakharov, and
  Carl Vondrick.
\newblock Zero-1-to-3: Zero-shot one image to 3d object.
\newblock In \emph{Proceedings of the IEEE/CVF International Conference on
  Computer Vision}, pages 9298--9309, 2023.

\bibitem[Lu et~al.(2023)Lu, Ball, and Parker-Holder]{lu2023synthetic}
Cong Lu, Philip~J Ball, and Jack Parker-Holder.
\newblock Synthetic experience replay.
\newblock \emph{arXiv preprint arXiv:2303.06614}, 2023.

\bibitem[Lugmayr et~al.(2022)Lugmayr, Danelljan, Romero, Yu, Timofte, and
  Van~Gool]{lugmayr2022repaint}
Andreas Lugmayr, Martin Danelljan, Andres Romero, Fisher Yu, Radu Timofte, and
  Luc Van~Gool.
\newblock Repaint: Inpainting using denoising diffusion probabilistic models.
\newblock In \emph{Proceedings of the IEEE/CVF Conference on Computer Vision
  and Pattern Recognition}, pages 11461--11471, 2022.

\bibitem[Lynch et~al.(2020)Lynch, Khansari, Xiao, Kumar, Tompson, Levine, and
  Sermanet]{lynch2020learning}
Corey Lynch, Mohi Khansari, Ted Xiao, Vikash Kumar, Jonathan Tompson, Sergey
  Levine, and Pierre Sermanet.
\newblock Learning latent plans from play.
\newblock In \emph{Conference on robot learning}, pages 1113--1132. PMLR, 2020.

\bibitem[Mandi et~al.(2022)Mandi, Bharadhwaj, Moens, Song, Rajeswaran, and
  Kumar]{mandi2022cacti}
Zhao Mandi, Homanga Bharadhwaj, Vincent Moens, Shuran Song, Aravind Rajeswaran,
  and Vikash Kumar.
\newblock Cacti: A framework for scalable multi-task multi-scene visual
  imitation learning.
\newblock \emph{arXiv preprint arXiv:2212.05711}, 2022.

\bibitem[Meng et~al.(2021)Meng, He, Song, Song, Wu, Zhu, and
  Ermon]{meng2021sdedit}
Chenlin Meng, Yutong He, Yang Song, Jiaming Song, Jiajun Wu, Jun-Yan Zhu, and
  Stefano Ermon.
\newblock Sdedit: Guided image synthesis and editing with stochastic
  differential equations.
\newblock \emph{arXiv preprint arXiv:2108.01073}, 2021.

\bibitem[Mildenhall et~al.(2021)Mildenhall, Srinivasan, Tancik, Barron,
  Ramamoorthi, and Ng]{mildenhall2021nerf}
Ben Mildenhall, Pratul~P Srinivasan, Matthew Tancik, Jonathan~T Barron, Ravi
  Ramamoorthi, and Ren Ng.
\newblock Nerf: Representing scenes as neural radiance fields for view
  synthesis.
\newblock \emph{Communications of the ACM}, 65\penalty0 (1):\penalty0 99--106,
  2021.

\bibitem[Pari et~al.(2021)Pari, Muhammad, Arunachalam, and
  Pinto]{pari2021surprising}
Jyothish Pari, Nur Muhammad, Sridhar~Pandian Arunachalam, and Lerrel Pinto.
\newblock The surprising effectiveness of representation learning for visual
  imitation.
\newblock \emph{arXiv preprint arXiv:2112.01511}, 2021.

\bibitem[Park and Wong(2022)]{park2022robust}
Jung~Yeon Park and Lawson L.~S. Wong.
\newblock Robust imitation of a few demonstrations with a backwards model.
\newblock In \emph{Advances in Neural Information Processing Systems
  (NeurIPS)}, 2022.

\bibitem[Pomerleau(1988)]{pomerleau1988alvinn}
Dean~A Pomerleau.
\newblock Alvinn: An autonomous land vehicle in a neural network.
\newblock In \emph{Advances in Neural Information Processing Systems
  (NeurIPS)}, volume~1, 1988.

\bibitem[Pomerleau(1991)]{pomerleau1991efficient}
Dean~A Pomerleau.
\newblock Efficient training of artificial neural networks for autonomous
  navigation.
\newblock \emph{Neural computation}, 3\penalty0 (1):\penalty0 88--97, 1991.

\bibitem[Ramesh et~al.(2022)Ramesh, Dhariwal, Nichol, Chu, and
  Chen]{ramesh2022hierarchical}
Aditya Ramesh, Prafulla Dhariwal, Alex Nichol, Casey Chu, and Mark Chen.
\newblock Hierarchical text-conditional image generation with clip latents.
\newblock \emph{arXiv preprint arXiv:2204.06125}, 1\penalty0 (2):\penalty0 3,
  2022.

\bibitem[Rombach et~al.(2022)Rombach, Blattmann, Lorenz, Esser, and
  Ommer]{rombach2021high}
Robin Rombach, Andreas Blattmann, Dominik Lorenz, Patrick Esser, and Bj{\"o}rn
  Ommer.
\newblock High-resolution image synthesis with latent diffusion models.
\newblock In \emph{Proceedings of the IEEE Conference on Computer Vision and
  Pattern Recognition (CVPR)}, 2022.

\bibitem[Ronneberger et~al.(2015)Ronneberger, Fischer, and
  Brox]{ronneberger2015u}
Olaf Ronneberger, Philipp Fischer, and Thomas Brox.
\newblock U-net: Convolutional networks for biomedical image segmentation.
\newblock In \emph{MICCAI}, pages 234--241. Springer, 2015.

\bibitem[Ross et~al.(2011)Ross, Gordon, and Bagnell]{ross2011reduction}
St{\'e}phane Ross, Geoffrey Gordon, and Drew Bagnell.
\newblock A reduction of imitation learning and structured prediction to
  no-regret online learning.
\newblock In \emph{Proceedings of the fourteenth international conference on
  artificial intelligence and statistics}, pages 627--635, 2011.

\bibitem[Saharia et~al.(2022)Saharia, Chan, Saxena, Li, Whang, Denton,
  Ghasemipour, Gontijo~Lopes, Karagol~Ayan, Salimans,
  et~al.]{saharia2022photorealistic}
Chitwan Saharia, William Chan, Saurabh Saxena, Lala Li, Jay Whang, Emily~L
  Denton, Kamyar Ghasemipour, Raphael Gontijo~Lopes, Burcu Karagol~Ayan, Tim
  Salimans, et~al.
\newblock Photorealistic text-to-image diffusion models with deep language
  understanding.
\newblock \emph{Advances in Neural Information Processing Systems},
  35:\penalty0 36479--36494, 2022.

\bibitem[Schaal(1996)]{schaal1996learning}
Stefan Schaal.
\newblock Learning from demonstration.
\newblock In \emph{Advances in Neural Information Processing Systems
  (NeurIPS)}, volume~9, 1996.

\bibitem[Sch\"{o}nberger and Frahm(2016)]{schoenberger2016sfm}
Johannes~Lutz Sch\"{o}nberger and Jan-Michael Frahm.
\newblock Structure-from-motion revisited.
\newblock In \emph{Conference on Computer Vision and Pattern Recognition
  (CVPR)}, 2016.

\bibitem[Sch\"{o}nberger et~al.(2016)Sch\"{o}nberger, Zheng, Pollefeys, and
  Frahm]{schoenberger2016mvs}
Johannes~Lutz Sch\"{o}nberger, Enliang Zheng, Marc Pollefeys, and Jan-Michael
  Frahm.
\newblock Pixelwise view selection for unstructured multi-view stereo.
\newblock In \emph{European Conference on Computer Vision (ECCV)}, 2016.

\bibitem[Shafiullah et~al.(2022)Shafiullah, Cui, Altanzaya, and
  Pinto]{shafiullah2022behavior}
Nur~Muhammad Shafiullah, Zichen Cui, Ariuntuya~Arty Altanzaya, and Lerrel
  Pinto.
\newblock Behavior transformers: Cloning $ k $ modes with one stone.
\newblock In \emph{Advances in Neural Information Processing Systems
  (NeurIPS)}, volume~35, pages 22955--22968, 2022.

\bibitem[Shafiullah et~al.(2023)Shafiullah, Rai, Etukuru, Liu, Misra, Chintala,
  and Pinto]{shafiullah2023bringing}
Nur Muhammad~Mahi Shafiullah, Anant Rai, Haritheja Etukuru, Yiqian Liu, Ishan
  Misra, Soumith Chintala, and Lerrel Pinto.
\newblock On bringing robots home.
\newblock \emph{arXiv preprint arXiv:2311.16098}, 2023.

\bibitem[Sohl-Dickstein et~al.(2015)Sohl-Dickstein, Weiss, Maheswaranathan, and
  Ganguli]{sohl2015deep}
Jascha Sohl-Dickstein, Eric Weiss, Niru Maheswaranathan, and Surya Ganguli.
\newblock Deep unsupervised learning using nonequilibrium thermodynamics.
\newblock In \emph{International conference on machine learning}, pages
  2256--2265. PMLR, 2015.

\bibitem[Song et~al.(2020{\natexlab{a}})Song, Zeng, Lee, and
  Funkhouser]{song2020grasping}
Shuran Song, Andy Zeng, Johnny Lee, and Thomas Funkhouser.
\newblock Grasping in the wild: Learning 6dof closed-loop grasping from
  low-cost demonstrations.
\newblock \emph{Robotics and Automation Letters}, 2020{\natexlab{a}}.

\bibitem[Song and Ermon(2019)]{song2019generative}
Yang Song and Stefano Ermon.
\newblock Generative modeling by estimating gradients of the data distribution.
\newblock In \emph{Advances in Neural Information Processing Systems
  (NeurIPS)}, volume~32, 2019.

\bibitem[Song et~al.(2020{\natexlab{b}})Song, Sohl-Dickstein, Kingma, Kumar,
  Ermon, and Poole]{song2020score}
Yang Song, Jascha Sohl-Dickstein, Diederik~P Kingma, Abhishek Kumar, Stefano
  Ermon, and Ben Poole.
\newblock Score-based generative modeling through stochastic differential
  equations.
\newblock \emph{arXiv preprint arXiv:2011.13456}, 2020{\natexlab{b}}.

\bibitem[Su et~al.(2021)Su, Yu, Zollh{\"o}fer, and Rhodin]{su2021nerf}
Shih-Yang Su, Frank Yu, Michael Zollh{\"o}fer, and Helge Rhodin.
\newblock A-nerf: Articulated neural radiance fields for learning human shape,
  appearance, and pose.
\newblock volume~34, pages 12278--12291, 2021.

\bibitem[Tancik et~al.(2023)Tancik, Weber, Ng, Li, Yi, Wang, Kristoffersen,
  Austin, Salahi, Ahuja, et~al.]{tancik2023nerfstudio}
Matthew Tancik, Ethan Weber, Evonne Ng, Ruilong Li, Brent Yi, Terrance Wang,
  Alexander Kristoffersen, Jake Austin, Kamyar Salahi, Abhik Ahuja, et~al.
\newblock Nerfstudio: A modular framework for neural radiance field
  development.
\newblock In \emph{Proceedings of the ACM SIGGRAPH Conference}, pages 1--12,
  2023.

\bibitem[Thrun(2002)]{thrun2002probabilistic}
Sebastian Thrun.
\newblock Probabilistic robotics.
\newblock \emph{Communications of the ACM}, 45\penalty0 (3):\penalty0 52--57,
  2002.

\bibitem[Tseng et~al.(2023)Tseng, Li, Kim, Alsisan, Huang, and
  Kopf]{tseng2023consistent}
Hung-Yu Tseng, Qinbo Li, Changil Kim, Suhib Alsisan, Jia-Bin Huang, and
  Johannes Kopf.
\newblock Consistent view synthesis with pose-guided diffusion models.
\newblock In \emph{Proceedings of the IEEE/CVF Conference on Computer Vision
  and Pattern Recognition}, pages 16773--16783, 2023.

\bibitem[Wang et~al.(2023)Wang, Li, Zhang, Driggs-Campbell, Wu, Fei-Fei, and
  Li]{wang2023d}
Yixuan Wang, Zhuoran Li, Mingtong Zhang, Katherine Driggs-Campbell, Jiajun Wu,
  Li~Fei-Fei, and Yunzhu Li.
\newblock $d^3$ fields: Dynamic 3d descriptor fields for zero-shot
  generalizable robotic manipulation.
\newblock \emph{arXiv preprint arXiv:2309.16118}, 2023.

\bibitem[Wang et~al.(2022)Wang, Hunt, and Zhou]{wang2022diffusion}
Zhendong Wang, Jonathan~J Hunt, and Mingyuan Zhou.
\newblock Diffusion policies as an expressive policy class for offline
  reinforcement learning.
\newblock \emph{arXiv preprint arXiv:2208.06193}, 2022.

\bibitem[Watson et~al.(2022)Watson, Chan, Martin-Brualla, Ho, Tagliasacchi, and
  Norouzi]{watson2022novel}
Daniel Watson, William Chan, Ricardo Martin-Brualla, Jonathan Ho, Andrea
  Tagliasacchi, and Mohammad Norouzi.
\newblock Novel view synthesis with diffusion models.
\newblock \emph{arXiv preprint arXiv:2210.04628}, 2022.

\bibitem[Xian et~al.(2023)Xian, Gkanatsios, Gervet, Ke, and
  Fragkiadaki]{xian2023chaineddiffuser}
Zhou Xian, Nikolaos Gkanatsios, Theophile Gervet, Tsung-Wei Ke, and Katerina
  Fragkiadaki.
\newblock Chaineddiffuser: Unifying trajectory diffusion and keypose prediction
  for robotic manipulation.
\newblock In \emph{Conference on Robot Learning}, pages 2323--2339. PMLR, 2023.

\bibitem[Xiao et~al.(2022)Xiao, Radosavovic, Darrell, and
  Malik]{xiao2022masked}
Tete Xiao, Ilija Radosavovic, Trevor Darrell, and Jitendra Malik.
\newblock Masked visual pre-training for motor control.
\newblock \emph{arXiv preprint arXiv:2203.06173}, 2022.

\bibitem[Yang et~al.(2023)Yang, Pavone, and Wang]{yang2023freenerf}
Jiawei Yang, Marco Pavone, and Yue Wang.
\newblock Freenerf: Improving few-shot neural rendering with free frequency
  regularization.
\newblock In \emph{Proceedings of the IEEE/CVF Conference on Computer Vision
  and Pattern Recognition}, pages 8254--8263, 2023.

\bibitem[Ye et~al.(2023)Ye, Li, Gupta, De~Mello, Birchfield, Song, Tulsiani,
  and Liu]{ye2023affordance}
Yufei Ye, Xueting Li, Abhinav Gupta, Shalini De~Mello, Stan Birchfield, Jiaming
  Song, Shubham Tulsiani, and Sifei Liu.
\newblock Affordance diffusion: Synthesizing hand-object interactions.
\newblock In \emph{Proceedings of the IEEE Conference on Computer Vision and
  Pattern Recognition (CVPR)}, pages 22479--22489, 2023.

\bibitem[Young et~al.(2021)Young, Gandhi, Tulsiani, Gupta, Abbeel, and
  Pinto]{young2021visual}
Sarah Young, Dhiraj Gandhi, Shubham Tulsiani, Abhinav Gupta, Pieter Abbeel, and
  Lerrel Pinto.
\newblock Visual imitation made easy.
\newblock In \emph{Proceedings of the Conference on Robot Learning (CoRL)},
  pages 1992--2005. PMLR, 2021.

\bibitem[Young et~al.(2022)Young, Pari, Abbeel, and Pinto]{young2022playful}
Sarah Young, Jyothish Pari, Pieter Abbeel, and Lerrel Pinto.
\newblock Playful interactions for representation learning.
\newblock In \emph{International Conference on Intelligent Robots and Systems},
  pages 992--999. IEEE, 2022.

\bibitem[Yu et~al.(2021)Yu, Ye, Tancik, and Kanazawa]{yu2021pixelnerf}
Alex Yu, Vickie Ye, Matthew Tancik, and Angjoo Kanazawa.
\newblock pixelnerf: Neural radiance fields from one or few images.
\newblock In \emph{Proceedings of the IEEE/CVF Conference on Computer Vision
  and Pattern Recognition}, pages 4578--4587, 2021.

\bibitem[Yu et~al.(2023{\natexlab{a}})Yu, Forghani, Derpanis, and
  Brubaker]{yu2023long}
Jason~J Yu, Fereshteh Forghani, Konstantinos~G Derpanis, and Marcus~A Brubaker.
\newblock Long-term photometric consistent novel view synthesis with diffusion
  models.
\newblock \emph{arXiv preprint arXiv:2304.10700}, 2023{\natexlab{a}}.

\bibitem[Yu et~al.(2023{\natexlab{b}})Yu, Xiao, Stone, Tompson, Brohan, Wang,
  Singh, Tan, Peralta, Ichter, et~al.]{yu2023scaling}
Tianhe Yu, Ted Xiao, Austin Stone, Jonathan Tompson, Anthony Brohan, Su~Wang,
  Jaspiar Singh, Clayton Tan, Jodilyn Peralta, Brian Ichter, et~al.
\newblock Scaling robot learning with semantically imagined experience.
\newblock \emph{arXiv preprint arXiv:2302.11550}, 2023{\natexlab{b}}.

\bibitem[Zhang et~al.(2023)Zhang, Rao, and Agrawala]{zhang2023adding}
Lvmin Zhang, Anyi Rao, and Maneesh Agrawala.
\newblock Adding conditional control to text-to-image diffusion models.
\newblock In \emph{Proceedings of the IEEE/CVF International Conference on
  Computer Vision}, pages 3836--3847, 2023.

\bibitem[Zhang et~al.(2018)Zhang, McCarthy, Jow, Lee, Chen, Goldberg, and
  Abbeel]{zhang2018deep}
Tianhao Zhang, Zoe McCarthy, Owen Jow, Dennis Lee, Xi~Chen, Ken Goldberg, and
  Pieter Abbeel.
\newblock Deep imitation learning for complex manipulation tasks from virtual
  reality teleoperation.
\newblock In \emph{Proceedings of the IEEE International Conference on Robotics
  and Automation (ICRA)}, pages 5628--5635, 2018.

\bibitem[Zhao et~al.(2023)Zhao, Kumar, Levine, and Finn]{zhao2023learning}
Tony~Z Zhao, Vikash Kumar, Sergey Levine, and Chelsea Finn.
\newblock Learning fine-grained bimanual manipulation with low-cost hardware.
\newblock \emph{arXiv preprint arXiv:2304.13705}, 2023.

\bibitem[Zhou et~al.(2023)Zhou, Kim, Wang, Florence, and Finn]{zhou2023nerf}
Allan Zhou, Moo~Jin Kim, Lirui Wang, Pete Florence, and Chelsea Finn.
\newblock Nerf in the palm of your hand: Corrective augmentation for robotics
  via novel-view synthesis.
\newblock In \emph{Proceedings of the IEEE/CVF Conference on Computer Vision
  and Pattern Recognition}, pages 17907--17917, 2023.

\bibitem[Zhou et~al.(2018)Zhou, Tucker, Flynn, Fyffe, and
  Snavely]{10.1145/3197517.3201323}
Tinghui Zhou, Richard Tucker, John Flynn, Graham Fyffe, and Noah Snavely.
\newblock Stereo magnification: learning view synthesis using multiplane
  images.
\newblock \emph{ACM Trans. Graph.}, 37\penalty0 (4), 2018.

\bibitem[Zhu et~al.(2018)Zhu, Wang, Merel, Rusu, Erez, Cabi, Tunyasuvunakool,
  Kram{\'a}r, Hadsell, de~Freitas, et~al.]{zhu2018reinforcement}
Yuke Zhu, Ziyu Wang, Josh Merel, Andrei Rusu, Tom Erez, Serkan Cabi, Saran
  Tunyasuvunakool, J{\'a}nos Kram{\'a}r, Raia Hadsell, Nando de~Freitas, et~al.
\newblock Reinforcement and imitation learning for diverse visuomotor skills.
\newblock \emph{arXiv preprint arXiv:1802.09564}, 2018.

\end{thebibliography}

\end{document}